\RequirePackage{fix-cm}
\documentclass{svjour4}       

\smartqed
 \usepackage{graphicx}
 \usepackage{amsmath}
 \usepackage{epsfig}
 \usepackage[ruled]{algorithm}
 \usepackage{algorithmic}

\newcommand{\ibmaphc}{\ensuremath{\mbox{IBMAP-HC}}}
\newcommand{\ibscore}{\ensuremath{\sigma} }
\newcommand{\HHCMN}{HHC-MN}

\newcommand{\moaMI}{\mbox{MOA} }
\newcommand{\moaIBMAP}{\mbox{MOA}' }

\newcommand{\set}[1]{\ensuremath{\mathbf{#1}}}

\newcommand{\PERP}{\mbox{\ensuremath{\perp\!\!\!\perp}}}
\newcommand{\NPERP}{\mbox{\ensuremath{\,\not\!\perp\!\!\!\perp}}}
\newcommand{\indep}[2]{\ensuremath{#1 \PERP #2}}
\newcommand{\dep}[2]{\ensuremath{#1 \NPERP #2}}

\newcommand{\eq}[1]{Eq.~(\ref{#1})}

\newcommand{\ci}[3]{\ensuremath{\langle \indep{#1}{#2} | #3} \rangle}
\newcommand{\cd}[3]{\ensuremath{ \langle \dep{#1}{#2} | #3} \rangle}

\newcommand{\blanket}[1]{\ensuremath{\set{B}_{#1}}}

\newcommand{\closure}{\ensuremath{\mathcal{C}}}

\newcommand{\argmin}[1]{\ensuremath{\operatorname*{arg\,min}\limits_{#1}}}

\begin{document}

\title{The IBMAP approach for Markov network structure learning}

\author{Federico Schl\"uter \and Facundo Bromberg \and Alejandro Edera }

\institute{F. Schl\"uter, F. Bromberg, A. Edera \at
	  Lab. DHARMa of Artificial Intelligence,  \\
	  Departamento de Sistemas de informaci\'on,  \\
	  Facultad Regional Mendoza, 
	  Universidad Tecnol\'ogica Nacional, Argentina. \\ 
              Tel.: +54-261-5244566 \\
              \email{\{federico.schluter,fbromberg,aedera\}@frm.utn.edu.ar} 
}

\date{Received: date / Accepted: date}

\maketitle

\begin{abstract}
  In this work we consider the problem of learning the structure of Markov networks from data.
  We present an approach for tackling this problem called IBMAP, together with an efficient
  instantiation of the approach: the \ibmaphc~algorithm, designed for avoiding
  important limitations of existing independence-based algorithms.
  These algorithms proceed by performing statistical independence tests
  on data, trusting completely the outcome of each test. In practice tests may be incorrect, 
  resulting in potential cascading errors and the consequent reduction in the quality 
  of the structures learned. IBMAP contemplates this uncertainty in the outcome 
  of the tests through a probabilistic maximum-a-posteriori approach.
  The approach is instantiated in the \ibmaphc~algorithm, a structure selection strategy 
  that performs a polynomial heuristic local search in the space of possible structures.
  We present an extensive empirical evaluation on synthetic and real data, 
  showing that our algorithm outperforms significantly the current independence-based algorithms, 
  in terms of data efficiency and quality of learned structures, 
  with equivalent computational complexities. 
  We also  show the performance of \ibmaphc~in a real-world application of knowledge discovery: EDAs,
  which are evolutionary algorithms that use structure learning on each generation for
  modeling  the distribution of populations. The experiments show that  
  when \ibmaphc~is used to learn the structure, EDAs improve the convergence to the optimum.

\end{abstract}


\section{Introduction} \label{sec:intro}

We present in this work the IBMAP (Independence-Based Maximum a Posteriori) approach 
for robust learning of Markov network structures from data, 
together with $\ibmaphc$, an efficient hill-climbing instantiation of the
approach.  Markov networks, together with Bayesian networks, belong to
the family of \textit{probabilistic graphical models} \cite{koller09}, a computational
framework for compactly representing joint probability distributions.
There is a large list of applications of graphical models 
in a wide range of fields, such as 
in the areas of computer vision and image analysis \cite{MCCALLUM03,Li2009},
computational biology \cite{friedman00},
biomedicine \cite{schmidts08,vanhaaren2013},
evolutionary computation \cite{larranagalozano2002,Alden07,moapaper}, among many others.

Probabilistic graphical models are composed by an undirected (Markov networks) or directed (Bayesian networks) graph $G$,
and a set of numerical parameters $\Theta$.  
Each node in the graph $G$ represents a random variable of the domain, 
and the edges encode conditional independences among them.  
For this reason, the graph $G$ is also called the \emph{independence structure} of the distribution.
The importance of these independences is that they factorize the joint
distribution over the domain variables into factors over subsets of
variables, resulting in important reductions in the space complexity 
for representing the distribution \cite{Hammersley_Clifford_1968}.
The structure can be obtained from the knowledge of a human expert, but commonly 
it is hard to obtain, and not always enough to design an accurate structure. 
An interesting problem that has attracted considerable attention 
is learning automatically the independence structure from categorical data 
drawn from an unknown probability distribution \cite{koller09,wainwright2008}.
However, this problem is known to be in general an NP-hard problem, 
since the number of structures grows super-exponentially \cite{Chickering96lns}.

For Markov network structure learning, 
there are two broad approaches mainly considered in the literature:
\emph{score-based} \cite{PietraPL97,MCCALLUM03,Lee+al:NIPS06,ganapathi2008}, 
and \emph{independence-based} (also known as constraint-based) algorithms \cite{Spirtes00,brombergmargaritis09b,margaritisBromberg09,Aliferis2010}. 
On the one hand, the score-based algorithms combine a measure of the goodness of fit of 
each structure to the data with a metric for the complexity of the structure;
for instance, to maximize the log-likelihood of the maximum likelihood parameters given the structure.
Recently, several efficient instantiations of this approach have been developed, 
such as \cite{ravikumar2010:l1,DavisAndDomingos2010:BottomUp,vanhaaren2012}.
On the other hand, the independence-based algorithms proceed by performing statistical independence tests on data, 
and based on the outcome of the tests discards
all structures inconsistent with the test. This approach 
is efficient, and correct under assumptions, but in practice
presents quality problems: one of the assumptions is
the correctness of independence tests, which may not be true in practice
when data are insufficient.
It is important to mention that both score-based and independence-based approaches
have been motivated by distinct learning goals.
According to the existent literature \cite{koller09}, score-based approaches are better suited for the density estimation goal, that is, 
tasks where inferences or predictions are required \cite{minka2005}.
In contrast, independence-based methods are better suited for other learning goals, such as
feature selection for classification, or knowledge discovery \cite{Spirtes00,Aliferis2010,Aliferis2010b}.

IBMAP follows the independence-based approach for learning the structure of a Markov network.
Our approach has been designed to be more robust when
the assumption of correctness of statistical tests is not valid. 
Instead of trusting the outcome of statistical tests on data, IBMAP 
considers explicitly the posterior probability of independences
given the data. As explained in detail later on, these posteriors of tests
are combined into the posterior of the whole structure (given the data), 
deciding on the output structure following the well-known
maximum-a-posteriori approach. This clearly circumvents the cascading 
error of traditional independence-based algorithms, 
as the true structure is no longer discarded on an incorrect
test, it only results in a lower posterior probability. 
With further tests, the posterior probability of the true structure may
increase again.

In order to evaluate the improvements in the quality of the structures produced by our approach,  we performed 
detailed and systematic experiments on both synthetic datasets and real-world datasets.
In all those cases we compared the structural errors of the structures learned by \ibmaphc~against 
those learned by representative state-of-the-art competitors: GSMN \cite{Bromberg06,brombergmargaritis09b}, 
and \HHCMN, a simple adaptation for Markov networks 
  of an independence-based structure learning algorithm for Bayesian networks,
  called HHC  \cite{Aliferis2010b}. We note that structural errors as quality measure
is the most appropriate for knowledge discovery algorithms such as 
those using the independence-based approach.

Additionally, we tested the performance of \ibmaphc~in a real world application: 
\textit{Estimation of Distribution algorithms} (EDAs) \cite{Muhlenbein96}.
These evolutionary algorithms are able to solve problems 
that are known to be hard for traditional Genetic Algorithms \cite{larranagalozano2002}.
EDAs are variations of the well-known evolutionary algorithms,
that replace the crossover and mutation stages for generating a new population of solutions
with a sampling of a probability distribution learned from the selected population.
Our experiment in EDAs is motivated by the fact that the quality of structure learning
is expected to influence the results of the optimization. 
This occurs because the structure learning step is made for each generation of the optimization,
and the populations are generated by sampling from the distribution learned. 
As more accurate the structure learned,
the more effective is the sampling for generating good solutions. 
In our experiment we tested \ibmaphc~in the Markovianity Optimization Algorithm (MOA) \cite{moapaper}, 
a state-of-the-art EDA, based on Markov network structure learning.
We show that MOA improves its convergence to the optimum when \ibmaphc~is used to learn the structure.

The rest of this work is organized as follows. 
Section~\ref{sec:stateoftheart} presents an overview of 
the independence-based learning approach and motivates our contribution. 
Section~\ref{sec:ibmap} presents the IBMAP approach, and Section~\ref{sec:algorithm} details our \ibmaphc~algorithm.
Section~\ref{sec:experiments1} shows our experiments on synthetic and real datasets, and
Section~\ref{sec:experiments2} shows our experiments on EDAs.
Finally, Section~\ref{sec:conclusions} summarizes this work, 
and poses several possible directions of future work.
The paper also includes two appendices at the end.

\section{Background \label{sec:stateoftheart}}
This section provides some background on Markov networks,
defines the problem of structure learning,
and motivates our independence-based approach.
Hereon, we use capital letters to denote single random variables, 
and the sets of variables in bold. 

A Markov network representing an underlying distribution $P(\set{V})$ 
over a domain of $n=|\set{V}|$ random variables $\set{V}$ 
consists in an undirected graph $G$, 
and a set of potential functions, 
defined by a set of numerical parameters $\Theta$.
The graph $G$ is a map of the conditional independences in $P(\set{V})$, 
and such independences can be read from the graph through
\emph{vertex separation}, considering that each pair of variables $(X,Y)$ 
are said to be vertex separated by a set of variables $\set{Z} \subseteq \set{V}\setminus \{X,Y\}$
when every path between $X$ and $Y$ in $G$ contains some node in $\set{Z}$ \cite{pearl88}. 

The distribution $P(\set{V})$ can be factorized into
a product of \emph{potential} functions $\phi_c(V_c)$
over the completely connected sub-graphs 
(a.k.a., \emph{cliques}) $V_c$ of its structure $G$ \cite{Hammersley_Clifford_1968}, that is,
\begin{equation*} \label{eq:Gibbs}
 P(\set{V}) = \frac{1}{Z}\prod_{c \in cliques(G)} \phi_c(V_c),
\end{equation*}
where \(Z\) is the \emph{partition function}, a constant that normalizes the product of potentials.
Such potential functions are parameterized by the set
of numerical parameters $\Theta$.

The problem of structure learning takes
as input a dataset $D$, which is assumed 
to be a representative sample of the underlying distribution $P(\set{V})$.
Commonly, $D$ is structured in a tabular format, 
with one column per random variable in the domain $\set{V}$, 
and one row per data point. 
The optimal solution of the problem is a perfect-map of $P(\set{V})$ \cite{pearl88}, 
that is, a structure that encodes all the dependences and all the independences 
present in $P(\set{V})$. 
The closer to a perfect-map, the better is the structure learned, 
and the better is the resulting Markov network for representing $P(\set{V})$.

Independence-based algorithms learn a  perfect-map
by performing a succession of statistical independence tests, 
discarding at each iteration all structures inconsistent with the outcome of the test, 
and deciding on the tests to perform next based on the outcomes learned so far. 
 
A statistical independence test is a statistic computed from $D$ 
for testing if two random variables $X$ and $Y$ are conditionally independent,
given some conditioning set of variables $\set{Z}$; 
where $X$, $Y$ and $\set{Z}$ are disjoint subsets of the domain $\set{V}$.
This \emph{independence assertion} is denoted by $\ci{X}{Y}{\set{Z}}$ 
(or $\cd{X}{Y}{\set{Z}}$ for the \emph{dependence assertion}).
The computational cost of a test is  proportional
to the number of rows in $D$, and the number of variables involved in the test.
Examples of independence tests used in practice are Mutual Information
\cite{covertomas91}, Pearson's $\chi^2$ and $G^2$ \cite{AGRESTI02}, 
the Bayesian test \cite{MARGARITIS05}, and for continuous Gaussian data 
the \emph{partial correlation} test \cite{Spirtes00}, among others.

There are several advantages of independence-based algorithms. 
First, they can learn the structure 
without interleaving the expensive task of parameter estimation, 
reaching sometimes polynomial complexities in the number of statistical tests performed. 
If the complete model is required, the
parameters can be estimated only once for the learned structure. 
Another important advantage of such algorithms is that they 
are guaranteed to learn the correct structure of the underlying distribution, as long as the following assumptions hold:
\emph{i)} \emph{graph-isomorphism}, i.e., the independences in the distribution can be encoded in an undirected graph; 
\emph{ii)} the underlying distribution is \emph{strictly positive}, i.e., $P(\set{V}) > 0$, for every assignment of $\set{V}$; and 
\emph{iii)} the outcomes of tests are correct, i.e., the independences learned are true in $P(\set{V})$.

Unfortunately, the third assumption is rarely true in practice,
as the number of contingency tables for which a statistic has to be computed grows exponentially
with the number of variables in the conditioning set of the test. Therefore, the effective
dataset from which the statistic is computed decreases exponentially in size, thus degrading
exponentially the quality of the statistics.  When tests outcome incorrect independences,
independence-based algorithms produce what is commonly called \emph{cascade errors} \cite{Spirtes00},
that not only discard the true underlying structure, but further confuse the algorithm
in the test to perform next.
Our approach tackles this main issue 
of independence-based algorithms 
by contemplating the uncertainty in the outcome 
  of the tests through a probabilistic maximum-a-posteriori approach.

\section{The independence-based MAP approach \label{sec:ibmap}}

We describe now the main contribution of this work: 
the IBMAP approach for Markov network structure learning. 
Our approach avoids the cascade errors of traditional independence-based algorithms
that trust completely the outcome of the statistical tests.
For this, the central idea of IBMAP is to pose the structure learning task as a maximum-a-posteriori problem, 
by computing the posterior probability of each possible structure given data.
Formally:
\begin{equation} \label{eq:maxG}
	  G^{\star} = \arg\max_{G}{\Pr(G \mid D)}.  
  \end{equation}
In our approach, the posterior $\Pr(G \mid D)$ is computed by combining the outcome of a set
of conditional independence assertions that determine the structure $G$.
We call this set the \emph{closure} of the structure.
The remaining of this section describes
how to use the closure for computing the posteriors $\Pr(G \mid D)$. 
Next, in Section \ref{sec:algorithm},
the \ibmaphc~algorithm is presented as an efficiently instantiation of the MAP optimization.

Let us first define formally the concept of a closure:
\begin{definition}[Closure] \label{def:closure}
Let $G$ be an undirected independence structure of a positive graph-isomorph distribution $P(\set{V})$.
The \emph{closure} of $G$ is a set of conditional independence assertions, $\closure(G) = \{c_i\}$, 
that are sufficient for determining $G$ completely.     
\end{definition}

Given the above definition, it is possible to replace $G$ by $\closure(G)$ in 
\eq{eq:maxG}, obtaining:
\begin{equation} 
	  G^{\star} = \arg\max_{G}{\Pr(\closure(G) \mid D)}.  
\end{equation}

The posterior of the closure given data can be seen 
as a joint probability distribution over its individual independence assertions, given data.
By applying the chain rule over the assertions in $\closure(G)$, we obtain:
\begin{equation} \label{eq:chainRule} 
    \Pr(\closure(G) \mid D) = \prod_{c_i \in \closure(G)} \Pr( c_i | c_1, \ldots, c_{i-1}, D ).
\end{equation}

To the best of the author's knowledge,
no method exists for computing exactly the
probabilities $\Pr( c_i | c_1, \ldots, c_{i-1}, D )$ of independence assertions conditioned 
on other independence assertions and data. 
A common approximation is to assume that all the independence assertions in the closure
\emph{are mutually independent}.
This assumption is made implicitly by all the independence-based Markov network structure learning algorithms
\cite{schluter2012survey}, because the statistical tests are used as a black box, 
only using data for deciding independence for each assertion $c_i$.
The result of applying this approximation to \eq{eq:chainRule} is the following expression: 
\[
	\Pr( \closure(G) \mid D) \approx \prod_{c_i \in \closure(G)} \Pr(c_i \mid D),
\]
which expressed in terms of logarithms to avoid underflow, results in the following expression that we call
the \emph{IB-score}:
\begin{equation}
 \ibscore(G) = \sum\limits_{c_i \in \closure(G)} \log  \Pr( c_i \mid D )   \label{eq:independenceApproximationLogs}.
\end{equation}
For computing the posteriors of each term $\log \Pr(c_i \mid D )$ 
we use the Bayesian test of conditional independence \cite{MARGARITIS05,margaritisBromberg09}.
Finally, since the log function is monotonic, the maximization of the IBMAP approach can be expressed as:
  \begin{equation} \label{eq:maxIBScore}
	  G^{\star} \approx \arg\max_{G}{~\ibscore(G)}.  
  \end{equation}
Although computable, this expression is still intractable,
as there are $2^{n \choose 2}$ possible undirected structures in the search space.

\section{The \ibmaphc~algorithm}
\label{sec:algorithm}

This section presents our structure learning algorithm \emph{\ibmaphc},
our instantiation of the IBMAP approach.
\ibmaphc~performs a heuristic hill-climbing search 
in the space of possible structures, thus its name.
We first give a high-level overview of the algorithm, 
and then we describe some specific aspects, 
such as the closure used for computing the IB-score,
the heuristic used for speeding-up the search, 
and the complexity of the overall algorithm.

\ibmaphc~searches the structure with maximum IB-score,
considering as neighboring structures all those structures that 
result from flipping only one edge (i.e., single-edge additions or deletions).
Algorithm~\ref{alg:ibmaphc} presents its pseudo-code. 
The algorithm has as input parameter a dataset $D$,
used for computing the statistical independence tests. 
The search starts at line~\ref{line:Gempty} by creating a structure $G$ with $n$ nodes 
(the number of variables in the domain) and no edges. 
Then, the IB-score of $G$ is computed in line~\ref{line:ibscoreg0} 
and saved in the variable \emph{current-score}.
The hill-climbing search starts in the loop of line~\ref{line:repeat}.
The loop iterates by calling the \emph{select-next-structure} function at line~\ref{line:flip}
to select the neighbor of $G$ with maximum score, which is saved in variable $G'$. 
Since the number of possible neighbor structures is $n \choose 2$,
this function is a heuristic for selecting the best neighbor, 
avoiding the expensive cost of computing the IB-score for all them. 
This is explained in detail in Section~\ref{sec:selectnextstructure}.
Then, in line~\ref{line:computeScore2} the score of the best neighbor is computed,
and saved in the variable \emph{neighbor-score}.
The algorithm stops when the neighbor proposed does not improve the current score,
a condition checked at line~\ref{line:endLoop}.
If the termination criterion is not reached, the variables $G$ and \emph{current-score}
are re-assigned by the variables $G'$ and \emph{neighbor-score} in lines~\ref{line:ascent1} and~\ref{line:ascent2}, 
and the process is repeated until a local optimum is found.

\begin{algorithm}[!ht]
\small
\caption{ \ibmaphc~(dataset $D$)}
\label{alg:ibmaphc}
\begin{algorithmic}[1]
\STATE $G \gets $ empty structure with $n$ nodes \scriptsize ~~~~~~~~~~\textit{// $n$ is the domain size} \small  \label{line:Gempty}
\STATE current-score $\gets \ibscore(G)$ \label{line:ibscoreg0} 
\STATE \textbf{repeat}   \label{line:repeat}
\STATE \quad $G'$ $\gets$ \emph{select-next-structure}$(G, \ibscore(G))$  \label{line:flip} \scriptsize ~~~~~\textit{// see Algorithm~\ref{alg:selectnextstructure} and Section~\ref{sec:selectnextstructure}} \small \\
\STATE \quad neighbor-score $\gets \ibscore(G')$ \label{line:computeScore2}  \scriptsize ~~~~~~~~~~~~~~~~~~~~~~~~~~~\textit{// see incremental computation in Section~\ref{sec:MB-closure}} \small
\STATE \quad \textbf{if} neighbor-score $\leq$ current-score \textbf{then}  \label{line:endLoop}
\STATE \quad \quad \textbf{return} $G$ \label{line:Return}  \scriptsize ~~~~~~~~~~~~~~~~~~~~~~~~~~~~~~~~~~~~~\textit{// local maximum reached} \small  
\STATE \quad \textbf{else} 
\STATE \quad \quad $G \gets G'$ \label{line:ascent1}
\STATE \quad \quad current-score $\gets$ neighbor-score \label{line:ascent2}  \scriptsize ~~~~~~~\textit{// an ascent in the hill-climbing search} \small  
\end{algorithmic}

\end{algorithm}

For computing the IB-score $\ibscore$ of the candidate structures 
(lines~\ref{line:ibscoreg0} and~\ref{line:computeScore2})
we define a closure called the \emph{Markov blanket closure},
presented in the next subsection.
This closure has been designed to determine a structure 
with a number of independence tests which is quadratic in the number of variables in the domain.


\subsection{Markov blanket closure}
\label{sec:MB-closure}
The \emph{Markov blanket closure} is a closure set that follows Definition~\ref{def:closure}.
This closure has been designed using the \emph{Markov blanket} 
of a domain variable $X$, denoted here $\blanket{X}$.
In terms of graphs, the Markov blanket of $X$ is defined
as the set of all the nodes connected by an edge to the node of $X$ in the structure \cite{pearl88,koller09}, 
i.e., its adjacency set. 
In terms of independences, this allows to consider
that $X$ is conditionally independent of all its non-adjacent variables
in the graph, given its Markov blanket.
By this property, we define the Markov blanket closure as a set of closures that can be computed independently,
one for each variable. Formally:
\begin{definition}[Markov blanket closure] \label{def:mbclosure}
	The \emph{Markov blanket closure} of a structure $G$ 
	is a set of assertions determined by the union of a set $\closure_X(G)$ of independence and dependence assertions 
	for each variable $X$ in the domain $\set{V}$, i.e., 
	    \begin{equation} \label{eq:closureMB}
		    \closure(G) = \bigcup\limits_{X \in \set{V}} \closure_X(G),
	    \end{equation}
	where each $\closure_X(G)$ is the union of two mutually exclusive sets of assertions:
	    \begin{eqnarray} \label{eq:closureMBX}
		    \closure_X(G) = &\Big\{& \cd{X}{Y}{\blanket{X} \!\setminus\! \{Y\}}  ~:~  Y \!\in\! \blanket{X}  \Big\} ~ \cup \nonumber \\
			    &\Big\{& \ci{X}{Y}{\blanket{X} }  ~:~ Y \!\notin\! \blanket{X}  \Big\},
	    \end{eqnarray}
	that is, for each neighbor of $X$ ($Y\in \blanket{X}$)
	add a conditional dependence assertion between both variables conditioning on $\blanket{X} \setminus \{Y\}$;
	and for each non-neighbor of $X$ ($Y\notin \blanket{X}$), add a conditional independence assertion between both variables  
	conditioned on $\blanket{X}$.  
\end{definition}
The following theorem states that the Markov blanket closure is indeed a closure, that is, 
it completely determines the structure $G$ used to construct it.
\begin{theorem} \label{thm:closure}
	Let $G$ be an undirected independence structure of a positive graph-isomorph distribution $P(\set{V})$.
	The \emph{Markov blanket closure} of $G$
	is a set of conditional independence assertions 
	that are sufficient for completely determining the structure $G$. 
	\begin{proof}
	  The formal proof of this theorem is presented in Appendix~\ref{appendix:mbclosure}. 
	\end{proof}
\end{theorem}
This closure contains $n\times(n-1)$ assertions,
a number which is quadratic in the size of the domain,
that is, $n-1$ assertions for each of the $n$ variables.
This allows to decompose
the computation of the IB-score of \eq{eq:independenceApproximationLogs}
in $n$ independent \emph{variable IB-scores}:
\begin{eqnarray}  \label{eq:decomposableIBScore}
\ibscore(G) = \sum\limits_{X\in \set{V}} \ibscore_{X}(G),
\end{eqnarray}
where $\ibscore_{X}(G) = \sum\limits_{c_i \in \closure_X(G)} \log  \Pr( c_i \mid D )$.
This decomposition permits to compute incrementally
the score of any neighbor structure $G'$, based on 
a previous computation of the score of a structure $G$.
Given that $G$ and $G'$ differs by an edge $(X,Y)$, 
the only blankets affected are $\blanket{X}$ and $\blanket{Y}$,
requiring to recompute only $\ibscore_X$ and $\ibscore_Y$,
and reusing the $(n-2)$ remaining variable IB-scores.
Consequently, the cost of computing $\ibscore(G')$ from $\ibscore(G)$ 
in line~\ref{line:computeScore2} of Algorithm~\ref{alg:ibmaphc}
is reduced from $n\times(n-1)$ to $2\times(n-1)$ tests, i.e.,
from $O(n^2)$ to $O(n)$ tests.

Finally, for convenience of the explanation of the \emph{select-next-structure} function in the next section,
let us further decompose \eq{eq:decomposableIBScore} 
considering that each variable IB-score $\ibscore_X(G)$ is composed by  
$(n-1)$ terms $\ibscore_{X,Y}(G)$, called \emph{pairwise IB-scores},
as follows: 
\begin{eqnarray}  \label{eq:decomposableIBScore2}
 \ibscore(G) = \sum\limits_{X\in \set{V}} ~~ \sum\limits_{Y \in \set{V} \setminus \{X\}}  \ibscore_{X,Y}(G).
\end{eqnarray}
According to \eq{eq:closureMBX}, each pairwise IB-score $\ibscore_{X,Y}$ is obtained 
by computing the following posterior from data:
\begin{equation} \label{eq:conditionals}
	\ibscore_{X,Y}(G) = \left\{
		\begin{array}{ll}
			\log \Pr(\cd{X}{Y}{\blanket{X}-\{Y\}} \mid D)   & \mbox{~~if $(X,Y)$ is an edge in $G$,}\\
			\log \Pr(\ci{X}{Y}{\blanket{X}} \mid D)   & \mbox{~~otherwise.} 
		\end{array} \right\}.
\end{equation}

The next section shows the heuristic used by the \emph{select-next-structure} function 
for reducing the computation time of finding the neighbor of a structure that maximizes the IB-score.

\subsection{Heuristic for selecting the best neighbor structure}
\label{sec:selectnextstructure}

The na\"ive procedure for selecting the neighbor structure with maximum score
would iterate over all the $n \choose 2$ neighbors that differ in one edge,
computing the IB-score of each one.
For each neighbor, it would be required to perform $n\times(n-1)$ statistical tests for computing
its IB-score using the Markov blanket closure, resulting in a total cost of $O(n^4)$ tests 
for each ascent in the hill-climbing search. 
By computing incrementally the IB-score of each neighbor,
the cost of each ascent still results in a cost of $2\times(n-1)$ statistical tests for each structure,
with a total cost of $O(n^3)$ tests for each ascent.
In order to reduce this expensive computation time, 
\ibmaphc~uses a heuristic that estimates 
the optimal neighbor without a single test computation, i.e., a cost of $O(1)$ test computations. 

\begin{algorithm}[!ht]
\small
\center
\caption{ select-next-structure~($G$, $\ibscore(G)$)}
\label{alg:selectnextstructure}
\begin{algorithmic}[1]
  \STATE $(X^*,Y^*)\gets$ $ \argmin{(X,Y)\in (\set{V}\times \set{V}), X\neq Y}$ $\ibscore_{X,Y}(G) + \ibscore_{Y,X}(G)$ \label{line:flip1}
  \STATE $G' \gets$ $G$ with $(X^*,Y^*)$ flipped \label{line:flip2}
  \RETURN $G'$
\end{algorithmic}
\end{algorithm}

The \emph{select-next-structure} function is shown
in Algorithm~\ref{alg:selectnextstructure}. It has as input parameter 
the current structure $G$ and its corresponding score $\ibscore(G)$,
which at this point is already computed.
The function first selects in line~\ref{line:flip1} the ``optimal'' pair $(X^*,Y^*)$
as the least accurate edge (or absence of edge) in the current structure $G$.
It can be done by representing $\ibscore(G)$ as a data structure  
which contains the $n\times(n-1)$ pairwise scores $\ibscore_{X,Y}(G)$,
using the decomposable form of \eq{eq:decomposableIBScore2}.
Then, the best neighbor $G'$ is constructed in line~\ref{line:flip2} as a copy of $G$ with the 
pair $(X^*,Y^*)$ flipped, and this is returned. 

To understand the minimization shown in line~\ref{line:flip1} of Algorithm~\ref{alg:selectnextstructure},
note that the number of neighbors differing by one edge
is the same than the number of different pairs of variables $(X,Y)$, i.e., $n\times(n-1)/2$ pairs.
From this point of view, \eq{eq:decomposableIBScore2} can be seen
as a sum of two pairwise IB-scores per each pair of variables, resulting in the following expression of the IB-score:
\begin{eqnarray}  \label{eq:decomposableIBScore3}
 \ibscore(G) = \sum\limits_{(X,Y)\in \set{V}\times\set{V}, X\neq Y} ~~ \ibscore_{X,Y}(G)+\ibscore_{Y,X}(G).
\end{eqnarray}
With this form of $\ibscore(G)$, it is clear that the minimization 
finds the pair $(X^*,Y^*)$ whose contribution to $\ibscore(G)$ is the smallest.
The assumption made by the heuristic is that the structure resulting from flipping $(X^*,Y^*)$ 
would be similar than maximizing the IB-score among the neighboring structures.

As explained in Section~\ref{sec:MB-closure}, 
for computing incrementally $\ibscore(G')$ from $\ibscore(G)$
only $\ibscore_X(G')$ and $\ibscore_Y(G')$ need to be recomputed. 
The approximation made in the minimization consists 
in assuming that $\ibscore_X(G') \approx \ibscore_{X,Y}(G')$, and 
$\ibscore_Y(G') \approx \ibscore_{Y,X}(G')$, 
ignoring the remaining terms $\ibscore_{X,W}(G')$ and $\ibscore_{Y,W}, W \subseteq \set{V}\setminus \{X,Y\}$. 
This is based in the fact that, from $G$ to $G'$,
it is expected a strong change in the terms $\ibscore_{X,Y}$ and $\ibscore_{Y,X}$,
since the posterior of dependence is used in one structure, and the posterior of independence is used in the other.
In contrast, the terms ignored are assumed to have a mild change,
because only the Markov blanket of $X$ and $Y$ has a change, and therefore
these assertions only vary in the conditioning set. 
The approximation is possible because the pairwise IB-scores corresponding 
to the flipped edge $\ibscore_{X,Y}(G')$ and $\ibscore_{X,Y}(G)$ are complementary in both structures $G$ and $G'$,
since the posterior of independence and the posterior of dependence sums $1$.
It allows to estimate $\ibscore_{X,Y}(G')$ from the same pairwise IB-score $\ibscore_{X,Y}(G)$,
without a single test computation.
This estimation is made implicitly by the minimization.

This heuristic assumes that the ignored terms should have a minimal impact 
in the search for the optimal neighbor. 
This is of course an approximation, and only empirical results may shed light on its effectiveness. 
In the worst case, the approximation would result in the selection of a sub-optimal neighbor. 
This, however, is not different from many optimization algorithms that follow sub-optimal paths (e.g.,
the well-known Metropolis-Hastings search algorithm
that may follow a sub-optimal neighbor according to its proposal distribution).
Given the complexity of the problem, the impact of this 
approximation can only be assessed empirically. 
Later experiments show that despite this approximation, our approach 
is useful for avoiding the cascade effect of traditional independence-based algorithms,
outperforming always the state-of-the-art algorithms when data are scarce.
Additionally, Appendix~\ref{appendix:landscape} presents empirical measurements of the 
landscape of the IB-score for several synthetic datasets, showing
that in most cases, our structure selection strategy finds nearly optimal scores.

\subsection{Complexity of \ibmaphc} \label{sec:complexity}

This section summarizes the 
resulting computational cost of the whole algorithm using the hill-climbing search,
the Markov blanket closure, and the \emph{select-next-structure} function.

To begin, the most expensive operation of the algorithm is 
the computation of the IB-score of the initial structure 
at line~\ref{line:Gempty} of Algorithm~\ref{alg:ibmaphc}, which is 
computed non-incrementally, using the $n\times(n-1)$ tests of the Markov blanket closure;
this is a cost of $O(n^2)$ tests.
Next, in the main loop of Algorithm~\ref{alg:ibmaphc}, 
calling the \emph{select-next-structure} function has a cost of $O(1)$,
and the incremental computation of $\ibscore(G')$ at line \ref{line:computeScore2} 
requires to compute $2\times(n-1)$ tests; this is a cost of $O(n)$.
Finally, denoting by $M$ the number of ascents until termination, 
the overall computational cost of the algorithm is $O(n^2 + Mn)$. 
Since $M$ can be obtained only empirically, 
the experimental section shows measurements of $M$ 
on different scenarios, proving empirically that 
$M$ is not a source of an extra degree in the complexity
because it grows sub-linearly with $n$, 
resulting in an overall computational complexity of $O(n^2)$ statistical tests. 


\section{Experimental results \label{sec:experiments1}}

This section describes several experiments on synthetic and real datasets for testing empirically the
robustness of our approach IBMAP, and the efficiency of our algorithm $\ibmaphc$.
We report a detailed and systematic experimental comparison  
between \ibmaphc~and state-of-the-art independence-based structure learning algorithms.
We show a comparison of the quality of structures learned by our solution, 
against the quality of structures learned by GSMN \cite{brombergmargaritis09b}, 
a state-of-the-art independence-based algorithm in terms of quality.
We introduce also a competitor called \HHCMN~as an adaptation for learning the structure of Markov networks 
of the HHC algorithm \cite{Aliferis2010b},
a state-of-the-art independence-based algorithm for learning Bayesian networks. 
For comparing all the algorithms on the same ground, we ran all of them using the
Bayesian test \cite{MARGARITIS05} as statistical independence test.

The GSMN algorithm learns a structure 
by finding the Markov blanket of each variable of the domain 
with the GS algorithm \cite{MARGARITIS00}, 
and then the solution structure is constructed by adding an edge between each variable and 
the variables found in its Markov blanket.
The GS algorithm learns the Markov blanket of a variable $X$ in two phases: 
the \emph{grow} and \emph{shrink} phases.
During the grow phase, the algorithm increases the tentative Markov blanket 
with every variable $Y$ that is found dependent on $X$,
conditioning on the currently tentative Markov blanket. 
At the end of this phase, the tentative Markov blanket contains all members
of the true Markov blanket, but potentially includes some false positives that are non-members.
These false positives are removed during the shrink phase, where variables found
independent of $X$ conditioned on the current Markov blanket are removed from this set.
At the end of this phase, the tentative Markov blanket matches the true Markov blanket, 
under the assumption of correctness of tests.
The computational complexity of this algorithm is $O(n^2)$ 
in the number of independence tests for discovering the structure.

The HHC algorithm learns the structure by learning the set of parents and children
(PC) of each variable through the interleaved HITON-PC with symmetry
correction algorithm \cite{Aliferis2003:HITON,Aliferis2010}. 
The pseudo-code of this algorithm can be seen at \cite{Aliferis2010} (Figure~6, page~192).
For learning the PC of a variable $X$, this algorithm starts with an empty candidate PC set,
ranking the variables by priority for inclusion in the candidate set by unconditional dependence with $X$,
and discarding the variables found unconditionally independent with $X$.
Then, the algorithm utilizes an inclusion heuristic function that 
accepts each variable into the candidate PC set. If
any variable inside the candidate set becomes independent with $X$ given some
subset of the candidate set, then the algorithm removes that variable from the candidate set and never
considers it again. The inclusion function and the elimination strategy 
are iterated interleaved until there are no more variables to examine for inclusion.
The complexity of the HITON-PC~is $O(n2^\tau)$, where $\tau$ is the largest size of the PC set found,
and the complexity of HHC is $O(n^22^\tau)$, because HITON-PC is executed for each variable of the domain.
For the case of Markov networks, the equivalent of the PC of a
variable is its neighbors, that is exactly its Markov blanket. It is
therefore expected that HITON-PC learns the Markov blanket of a Markov network, and thus
it can be used as part of HHC to learn the undirected structure.
This fact is not proven analytically here, 
but confirmed empirically for all the cases considered in this section. 
To get a Markov network learning algorithm we simply omit the final step of HHC that
orients the edges to obtain the Markov blanket from the PC set, denoting the resulting algorithm by \HHCMN.  

The three following subsections describe our experiments over synthetic 
(Sections~\ref{sec:randomexperiments} and \ref{sec:isingexperiments}) and real datasets (Section~\ref{sec:benchmarkexperiments}).

\subsection{Synthetic data experiments: random underlying structures \label{sec:randomexperiments}}

A first set of experiments was conducted on synthetic datasets, 
generated by using a Gibbs sampler on randomly generated Markov networks (structure plus parameters). 
This allows a systematic and controlled study, 
and provides datasets with known underlying structures 
to control the complexity of the problem,
and to better assess the quality of the structures learned by each algorithm.

For measuring the structural errors of the structures learned,
we report the \emph{Hamming distance} between the learned structure and the underlying one, i.e., the
sum of false positive and false negative edges of the learned structure. 
Another quality measure that we use in this work for assessing the structures learned, 
is the well known F-measure, a harmonic mean of precision and recall quality measures, 
commonly used in the information retrieval community.
Precision indicates how good was the algorithm in learning correct independences 
(that is, the relation between the true independences that were found, 
over all independences found by the algorithm).
Instead, recall indicates how good was the algorithm in learning independences, 
but over all the correct independences present in the real structure 
(that is, the relation between the correct independences that were found, 
over the total of independences in the underlying structure).
Then, the F-measure is computed as follows:
\begin{equation*} \label{eq:fmeasure}
	\mbox{F-measure}=\frac{2 \times precision \times recall}{precision + recall}.
\end{equation*}

Additionally, at the end of this section, we show the runtime 
of our experiments, in order to discuss the computational complexities of the competitor algorithms.

\begin{figure}[!ht]
\center
\scriptsize
Random structures: Hamming distance results.
\normalsize

 \includegraphics[width=1.08\textwidth]{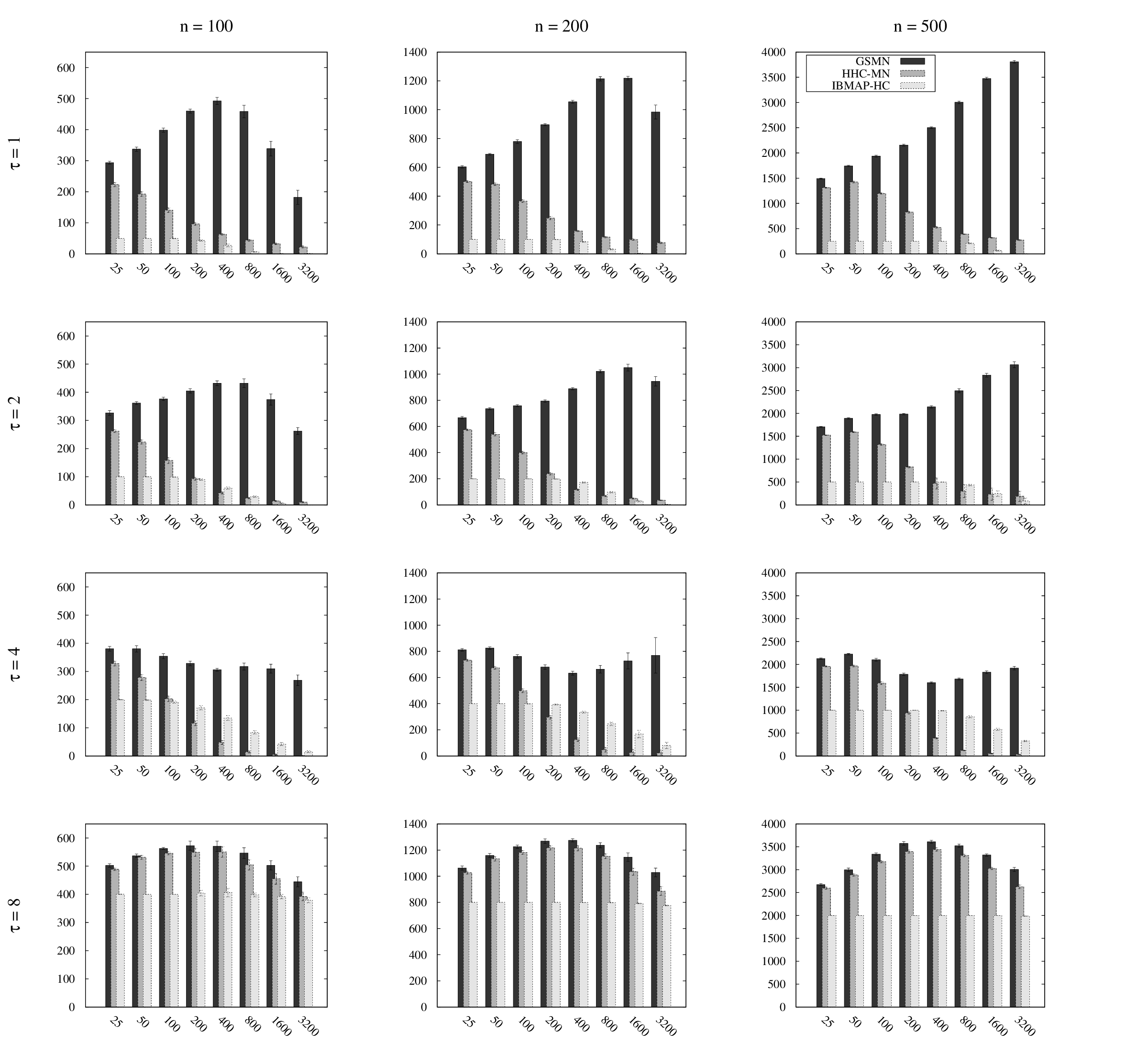} 
  \caption{ \small Mean and standard deviation over $10$ repetitions of the Hamming distance 
  of the models learned by algorithms GSMN~(black bars), \HHCMN~(gray bars), and \ibmaphc~(light gray bars)
  for increasing sizes of random synthetic datasets, domain sizes $n=100$ (first column), $n=200$ (second column), and
  $n=500$ (third column), and $\tau \in \{1,2,4,8\}$ in the rows. \label{fig:resultsHD} }
\end{figure}

The synthetic random Markov networks were generated for domains of $n \in \{100, 200, 500\}$ binary variables. 
For each domain size, $10$ random networks were generated for increasing
connectivities $\tau \in \{1,2,4,8\}$, by considering as edges the first $n\tau/2$
variable pairs of a random permutation of the set of all variable pairs. 
It is worth mentioning that with increasing values of $\tau$, it is increasingly difficult to learn the structure. 
Given these Markov networks, we report the quality of structures learned by GSMN, \HHCMN, and \ibmaphc~using 
portions of each dataset with increasing number of datapoints 
$D \in \{25,50,100,200,400,800,1600,3200\}$, for each $(n,\tau)$ combination. 

The independence structure determines the factorization of the distribution into 
potential functions over subset of variables, one per clique in the structure.
To determine a complete model
we must determine the numerical parameters that quantify these potential functions.
For the datasets generated to correctly and strongly represent the direct dependencies encoded
by the edges, we considered in these experiments pairwise cliques for the factorization of the models, 
that is, two-variable factors
$\phi(X,Y)$ for each edge in the random structure generated, and set the numerical parameters
so that the correlation between them is strong. For that, we forced the parameters to
result in a  log-odds ratio of each pairwise factor
$\varepsilon_{X,Y} = \log \left( \frac{\phi(X=0,Y=0) \phi(X=1,Y=1)}{\phi(X=0,Y=1) \phi(X=1,Y=0)} \right)$ 
to be equal to $1.0$ for all edges (see \cite{AGRESTI02}).
This results in an equation over the values of the potential function with 4 unknowns. 
We then randomly chose 3 parameters in the range [0, 1], and solved for the remaining one. 

\begin{figure}[!ht]
\center
\scriptsize
Random structures: F-measure results
\normalsize

 \includegraphics[width=1.08\textwidth]{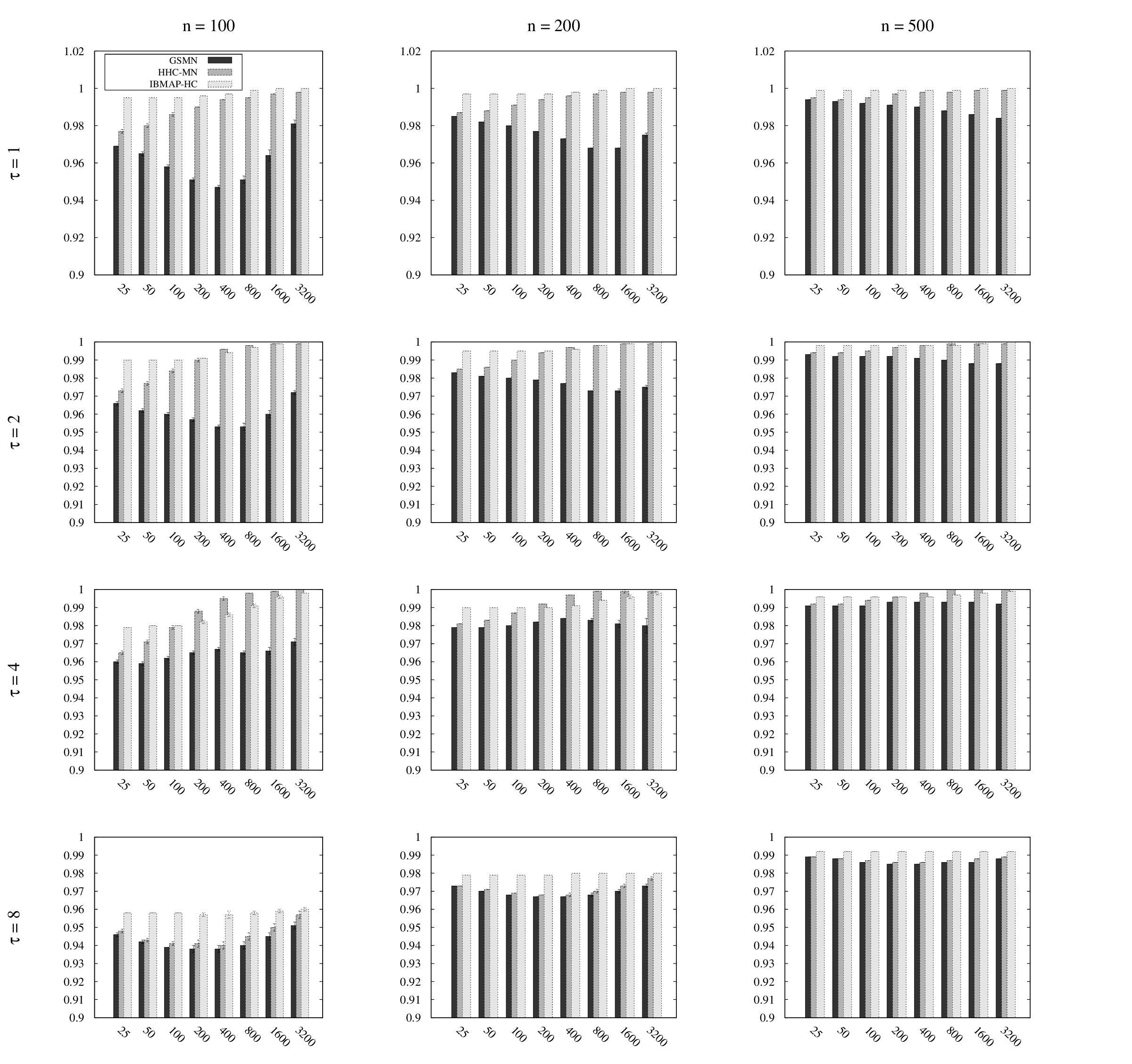} 
  \caption{ \small Mean and standard deviation over $10$ repetitions of the F-measure
  of the models learned by algorithms GSMN~(black bars), \HHCMN~(gray bars), and \ibmaphc~(light gray bars)
  for increasing dataset sizes of random synthetic datasets, domain sizes $n=100$ (first column), $n=200$ (second column), and
  $n=500$ (third column), and $\tau \in \{1,2,4,8\}$ in the rows. \label{fig:resultsFMeasure}}
\end{figure}

Figures~\ref{fig:resultsHD} and~\ref{fig:resultsFMeasure} show the mean values and
standard deviations over the ten repetitions 
of the Hamming distances and F-measure for the structures learned by the algorithms considered, respectively.  
The plots are ordered by columns for different $n$ values, 
and by rows for different $\tau$ values.
As expected, the results show that for all the algorithms, 
the more complex the underlying structure (determined by $n$ and $\tau$),
the larger is the number of structural errors for any value of $D$ used.
It can be seen that for any algorithm and for any fixed value of $D$,
the amount of errors grows with $n$ (different columns),
and also it grows with $\tau$ (different rows).

Since GSMN and \HHCMN~follow the traditional independence-based approach, 
it is expected for them to obtain very good qualities when data are sufficient, 
i.e., those cases with larger values of $D$ and lower values of $\tau$.
The figures show clearly that both, \ibmaphc~and \HHCMN~ always learn structures with qualities significantly better 
(lower Hamming distance, and higher F-measure) than that of GSMN. 
For all the cases of $n$ and $\tau$, GSMN has the slowest convergence in $D$ to reduce the structural errors. 
This is because, for the selected domain sizes, 
GSMN tend to add many false positives in the grow phase,
and then the shrink phase require to perform tests that contains many variables, 
i.e., that are not reliable. It produces numerous cascade errors.

In the case of \HHCMN, it can be seen that the structural errors are reduced significantly
with respect to GSMN. These improvements are obtained by the use of 
its elimination strategy, as well as the interleaving 
of the inclusion heuristic function with the elimination strategy.
When compared to \ibmaphc, the latter always outperforms \HHCMN~in terms of structural errors, 
except in the following specific cases:
\begin{itemize}
\item[$\bullet$] $\tau=2, n \in \{100,200,500\}, D \in \{400,800\}$
\item[$\bullet$] $\tau=4, n \in \{100,200,500\}, D \ge 200$.    
\end{itemize}
In the above cases the data seem to be sufficient 
for \HHCMN~to improve the quality of our algorithm \ibmaphc.
This is because for $\tau < 8$ the underlying structures have not a dense topology, 
and the elimination strategy results to be very efficient.
In contrast, for the case of $\tau=8$, 
the data are not sufficient for \HHCMN~to work as well,
due to the exponential size of tests required in the elimination strategy.
In this extreme case, the conditioning sets are at average of $8$ variables,
and in those cases the tests require larger amounts of data to be reliable.

In general, the figures confirm that \ibmaphc~always outperforms 
significantly the competitors when data are scarce ($D \leq 100$).
This confirm our hypothesis that the probabilistic approach of IBMAP
avoids the cascade effect of traditional independence-based algorithms.
Also, when the data are sufficient ($D>100$) the qualities obtained are very competitive.

\begin{figure}[!ht]
\center
\scriptsize
Random structures: Runtime results (in milliseconds)
\normalsize

 \includegraphics[width=1.08\textwidth]{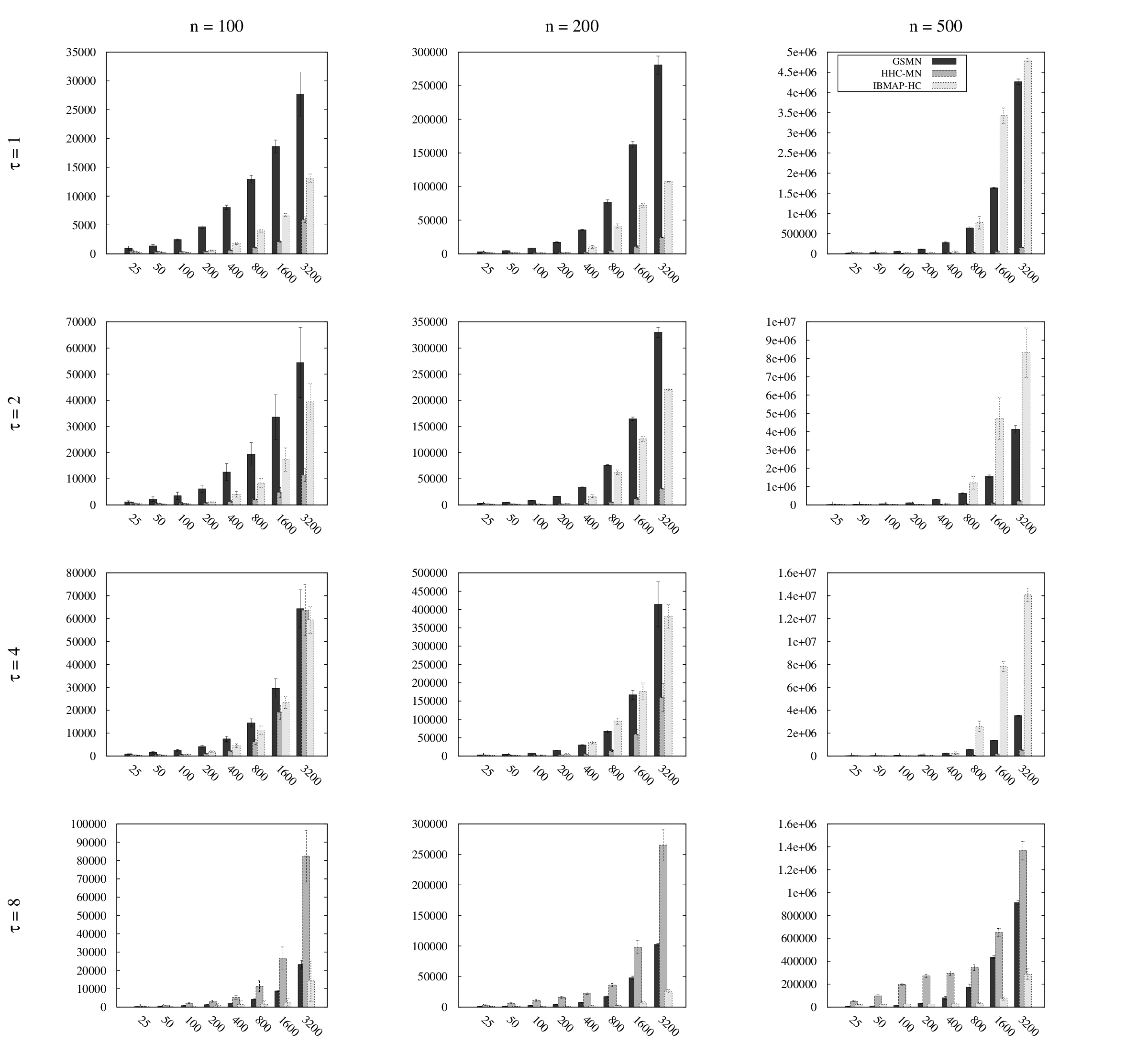} 
  \caption{ \small Mean and standard deviation over $10$ repetitions of the runtime
  required by algorithms GSMN~(black bars), \HHCMN~(gray bars), and \ibmaphc~(light gray bars)
  for increasing dataset sizes of random synthetic datasets,  domain sizes $n=100$ (first column), $n=200$ (second column), and
  $n=500$ (third column), and $\tau \in \{1,2,4,8\}$ in the rows. \label{fig:resultsRuntimes}}
\end{figure}

Figure~\ref{fig:resultsRuntimes} shows the corresponding running times of the same experiment, expressed in milliseconds.
To give the times more meaning, take into account that all our experiments were 
performed on an AMD Athlon(tm), with 3.0 GHz and 4 Gb of main memory.
Our results show clearly that GSMN is the more expensive algorithm in all the cases of $\tau \in \{1,2,4\}$.
This is because it tend to add many false positives in the grow phase,
and then the shrink phase require to perform tests that contains many variables, 
which is a source of extra computational cost. 
There are some extreme cases where \ibmaphc~is more expensive than GSMN, such as $n=500$, $\tau\in\{1,2,4\}$, and $D\geq 800$.
In those cases, the hill-climbing search of \ibmaphc~seem to be
the more expensive alternative.

\HHCMN~is the algorithm that requires lowest computation time 
for the cases of $\tau \in \{1,2,4\}$, and $D\geq 200$.
This is because the inclusion heuristic interleaved with the elimination strategy 
is really effective when the underlying structure has a low value of $\tau$, 
and $D$ is sufficiently large to obtain more reliable tests.
In these situations, the algorithm converge to the termination criterion quickly.
Instead, in the case of $\tau=8$ (last row), \HHCMN~is the most expensive algorithm.
This is due to the exponential cost of the elimination strategy, 
that performs a test for all the subsets of the current conditioning set, which in this case is $8$, on average.

\begin{figure}[!ht]
\center
\begin{tabular}{c}
\includegraphics[width=8cm]{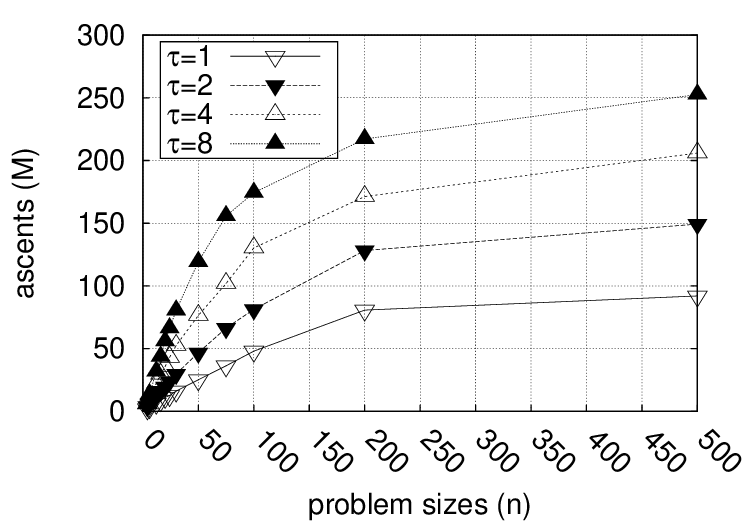} 
\end{tabular}
\vspace{-0.3cm}
\caption{ \small \label{fig:ibmaphc-space-complexity} Measurements in the number of ascents $M$ (Y-axis)
in the hill-climbing search of $\ibmaphc$ for increasing values of $n$ (X-axis), 
and $\tau\in\{1,2,4,8\}$, $D=1000$.}
\end{figure}

To conclude this section, we show an additional experiment to confirm empirically 
that \ibmaphc~achieves polynomial time complexities 
with the number of random variables in the domain, as stated in Section~\ref{sec:complexity}. 
This is shown by Figure~\ref{fig:ibmaphc-space-complexity}, that presents measurements of $M$ 
(number of ascents in the hill-climbing search) for increasing problem sizes $n$.
Such results were obtained for datasets generated in the same way as the previous experiments.
The figure shows the average values of $M$ over ten repetitions, for problems with increasing values 
of $n \in \{4,12,16,20,24,30,50,75,100,200,500\}$ in the X-axis, 
$D=1000$, and a line for each $\tau\in\{1,2,4,8\}$, indicating that $M$ (Y-axis) grows sub-linearly.
We omit results for different $D$ values because they are similar.

\subsection{Synthetic data experiments: Ising models \label{sec:isingexperiments}}

A second set of experiments over synthetic datasets were conducted 
over underlying structures with a different topology: the Ising spin glasses models, 
that are mathematical models of ferro-magnetism in statistical mechanics,
also used in the last decades in many other domains, such as computer vision applications \cite{Li2009}. 
Using such models as underlying structure, 
ten datasets were generated
for random Ising models with $n \in \{100, 200,500,750\}$ binary variables. 

\begin{figure}[!ht]
\center
\scriptsize
Ising models
\normalsize
 \includegraphics[width=\textwidth]{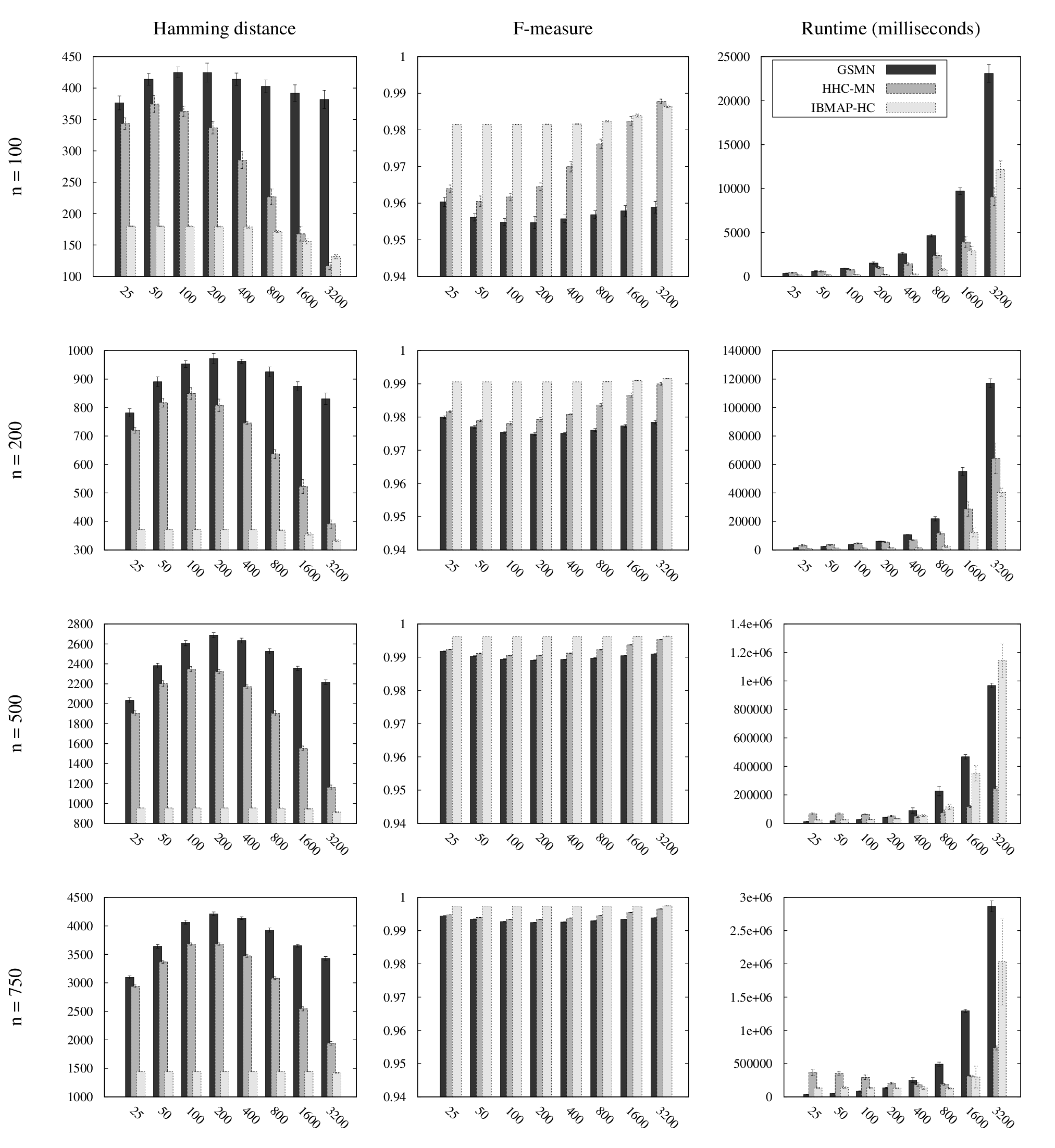} 
  \caption{ \small Mean and standard deviation over $10$ repetitions of the Hamming distance (first column), 
  F-measure (second column) and runtime (third column) of algorithms 
  GSMN~(black bars), \HHCMN~(gray bars), and \ibmaphc~(light gray bars)
  for increasing dataset sizes of Ising synthetic datasets, and domain sizes $n \in \{ 100, 200, 500, 750\}$ in the rows. \label{fig:resultsIsing}}
\end{figure}

Figure~\ref{fig:resultsIsing} shows the results for ten different random repetitions.
The graphs in this figure are ordered by rows for different $n$ values, 
and showing the mean value and standard deviation of the Hamming distance, the F-measure and the runtime 
in the first, second and third columns, respectively.
These figures show clearly that both, \ibmaphc~and \HHCMN~always learn structures with 
lower Hamming distance, and higher F-measure than that of GSMN (first and second column). 
In all the cases, the GSMN algorithm has the slowest convergence in $D$ to reduce the structural errors
among the three algorithms. 
With respect to \HHCMN, it can be seen that it has always lower structural quality than that of \ibmaphc,
except in the specific case of $n=100$, $D=3200$, 
where the data seem to be sufficient 
for \HHCMN~to improve the quality of \ibmaphc.
In general, the figures confirm that \ibmaphc~outperforms 
significantly the competitors in terms of quality. 
This also confirm our hypothesis that the probabilistic approach of IBMAP
avoids the cascade effect of the traditional independence-based algorithms.

With regard to the computational complexity results (third column), 
Figure~\ref{fig:resultsRuntimes} shows the corresponding running times, expressed in milliseconds.
The computer used for running these experiments was the same described in the previous section. 
These results show clearly that GSMN is the more expensive algorithm for all the cases,
except in the specific cases:
\begin{itemize}
\item[$\bullet$] $n \in \{200,500,750\}, D \leq 100$, where \HHCMN~is the more expensive;
\item[$\bullet$] $n \in \{500\}, D = 3200$, where \ibmaphc~is the more expensive.    
\end{itemize}
For the rest of the cases, \ibmaphc~has the better runtime, 
except in the following cases, where \HHCMN~has the better runtime:
\begin{itemize}
\item[$\bullet$] $n \in \{100,750\}, D = 3200$;
\item[$\bullet$] $n \in \{500\}, D \geq 800$.    
\end{itemize}

The analysis of these runtime results are similar than the runtime analysis of the previous section,
with GSMN with an expensive cost, due to the large amount of expensive tests (many false positives in the conditioning set),
\HHCMN~with a very good performance when data are sufficient, and \ibmaphc~with the best performance 
when data are not sufficient ($D<200$).

\subsection{Benchmark datasets experiments \label{sec:benchmarkexperiments}}

In this section we show our experiments on real-world benchmark datasets, 
obtained from the UCI Repositories of machine learning
\cite{Asuncion+Newman:2007} and KDD datasets \cite{Hettich+Bay:1999}.               
Since the underlying network is unknown in these datasets,
it is not possible to compute neither the Hamming distance nor the F-measure. 
Instead, we utilize the \emph{accuracy}, a quality measure that counts
the number of conditional independences present in data, which are correctly 
encoded by the structure learned. This measure was used 
for the same purpose in other related works \cite{brombergmargaritis09b,margaritisBromberg09,BrombMarg09}.
In contrast with other measures that evaluate 
the density of the complete probability distribution (e.g. the Conditional Marginal Log-Likelihood),
the accuracy is better suited for the goal of learning of this work (knowledge discovery)
because it evaluate specifically structural errors. 

The accuracy is defined as a normalized measure for counting the 
number of matches in a comparison of the independence queries
that hold in a \emph{test set}, and also hold in the structure learned from a \emph{training set}. 
The conditional independences are read from the learned structure by vertex separation (see Section~\ref{sec:stateoftheart}).
If $\mathcal{T}$ denotes the set of all possible conditional independence 
queries over the set of domain variables $\set{V}$,   
it is checked for how many queries $t \in \mathcal{T}$, $t$ is independent (or dependent) 
in both the test set, and the learned structure from the training set. 
Then, the number of matches is normalized by $|\mathcal{T}|$.
Unfortunately, the size of $\mathcal{T}$ is exponential, so the
approximated accuracy is computed over a randomly sampled subset $\widehat{\mathcal{T}}$, 
uniformly distributed for each possible conditioning set size.
In our experiments we used $| \widehat{\mathcal{T}}| = 100 \times {n \choose 2}$, 
i.e., a hundred of conditional independence queries per conditioning set size.

We conducted our experiment using $19$ real-world datasets, listed in Table~\ref{table:real}, column one.
The datasets are sorted by domain size ($n$) in the second column.  
For each dataset $D$, we shuffled the data and then divided it
into a training set for learning the structure ($75\%$), and a test set for computing the accuracy ($25\%$). 
The table also shows information about the number of attributes (second column), 
and the number of datapoints available in the train and test sets (third and fourth column).
For each dataset we used the train set as input to the GSMN, \HHCMN, and \ibmaphc~algorithms,
and the accuracy obtained for the structure learned for each algorithm
is shown in the fifth, sixth and seventh columns, respectively.
For each dataset, the best performance among the three algorithms is indicated in bold. 
These results show that in $10$ of $19$ datasets 
\ibmaphc~resulted in better accuracy, $6$ cases resulted in ties ($2$ with GSMN, $1$ with \HHCMN, and $3$ with both),
and for the remaining cases, the best results are obtained by \HHCMN ($2$ cases) and GSMN ($1$ case).
The cases where \ibmaphc~always outperforms it competitors are 
those with $n\geq 16$. In those cases, data seem to be scarce (see the third column).
That is consistent with our results in synthetic datasets, 
where IBMAP-HC outperforms always its competitors when data are scarce. 

\begin{table}[!ht]
\center

		      \begin{tabular}{|c|c|c|c|c|c|c|}
			    \hline 
			                     &              & Train             & Test & \multicolumn{3}{|c|}{\textbf{accuracy}} \\ \cline{5-7}
			    \textbf{Dataset} & \textbf{$n$} & \textbf{$D$} & \textbf{$D$} & \textbf{GSMN} & \textbf{\HHCMN} & \textbf{\ibmaphc} \\ \hline
			    baloons & 5 & 14 & 5 & \textbf{0.950} & 0.897 & \textbf{0.950} \\
			    balance-scale & 5 & 468 & 156 & \textbf{0.516} & \textbf{0.516} & \textbf{0.516} \\
			    iris & 5 & 112 & 37 & 0.695 & \textbf{0.742} & 0.736 \\
			    lenses & 5 & 17 & 6 & \textbf{0.881} & 0.875 & \textbf{0.881} \\
			    hayes-roth & 6 & 98 & 33 & \textbf{0.516} & \textbf{0.516} & \textbf{0.516} \\
			    car & 7 & 1295 & 432 & 0.629 & 0.641 & \textbf{0.703} \\
			    monks-1 & 7 & 416 & 139 & \textbf{0.905} & \textbf{0.905} & \textbf{0.905} \\
			    nursery & 9 & 9719 & 3240 & 0.392 & 0.415 & \textbf{0.649} \\
			    ecoli & 9 & 251 & 84 & 0.523 & 0.591 & \textbf{0.694} \\
			    machine & 10 & 156 & 52 & 0.590 & 0.567 & \textbf{0.679} \\
			    cmc & 10 & 1104 & 368 & \textbf{0.759} & 0.711 & 0.726 \\
			    tic-tac-toe & 10 & 718 &  239 & 0.671 & \textbf{0.684} & 0.498 \\
			    echocardiogram & 13 & 45 & 15 & 0.696 & \textbf{0.745} & \textbf{0.745} \\
			    crx & 16 & 489 & 163 & 0.578 & 0.593 & \textbf{0.609} \\
			    hepatitis & 20 & 59 & 20 & 0.496 & 0.633 & \textbf{0.796} \\
			    imports-85 & 25 & 144 & 28 & 0.368 & 0.377 & \textbf{0.596} \\
			    flag & 29 & 145 & 48 & 0.446 & 0.451 & \textbf{0.803} \\
			    dermatology & 35 & 268 & 53 & 0.234 & 0.265 & \textbf{0.754} \\
			    bands & 38 & 207 & 69 & 0.399 & 0.408 & \textbf{0.546} \\
\hline
			\end{tabular}

	      \caption{ \small
Accuracy for several benchmark data sets. 
The structure is learned using a subsample called train set, and the accuracy is computed using the test set.
For each evaluation measure, the best performance is indicated in bold.
\label{table:real}}
\end{table}

\section{\ibmaphc~for Estimation of Distribution Algorithms \label{sec:experiments2}}

In contrast to benchmark datasets that comes from arbitrary applications, 
we present now results of evaluating \ibmaphc~in a real world application of knowledge-discovery:
the \textit{Estimation of Distribution algorithms} (EDAs) \cite{Muhlenbein96,larranagalozano2002}.
These are variations of the well-known evolutionary algorithms,
that perform the same \emph{selection} and \emph{variation} stages, but replace
the \emph{crossover} and \emph{mutation} stages with the \emph{estimation} and \emph{sampling} 
in the task of generating a new population.
The former stage \emph{estimate} a probability
distribution from the current population,
generating the next population by \emph{sampling}
from it (thus their name).
In the \emph{estimation} stage, EDAs estimate the probability
distribution from the dataset corresponding to the current population.
This is because they associate each gene to a random variable,
each individual to a joint assignment of these variables,
and the selected population to a sample of the distribution.
The rationale for replacing crossover methods with estimation is that
by estimating the distribution
from the selected individuals, that is, those best fitted,
the sampling stage would produce novel, yet well-fitted individuals.

Recently, several Markov network based EDAs has been proposed to model the distribution 
of populations \cite{Santana:2005:kikuchi,Alden07,Shakya_McCall_2007,moapaper}.
As a test-bed we considered the \emph{Markovianity Optimization Algorithm} (MOA) \cite{moapaper}.
This is a state-of-the-art MN-based EDA that learns the Markov network structure from the population
using an efficient structure learning algorithm
based on mutual information (MI), a simple independence-based 
structure learning algorithm, described in detail in the same work, 
and designed specifically for MOA.
The sampling in MOA is conducted through a variation of a Gibbs sampler that
requires only the structure of the model, avoiding the need to learn
the model parameters.
The implementation of MI in MOA takes advantage of experts information
indicating the maximum number of neighbor variables that a variable can have,
denoted here $k$.
We tested MI for different values of $k$
(results not shown here), observing great sensitivity of MI
to its value. Our algorithm \ibmaphc~does not use such a parameter. 
In the experiments below we set the value of $k$ for MI
to be the closest to the true value, resulting in
the best possible performance of MI, i.e., the strongest
competitor for $\ibmaphc$.

\begin{table*}[!ht]
\center
  \begin{tabular}{cc|c||c|c||}
   \\ \hline
  \multicolumn{1}{||c||}{}& \multicolumn{2}{c||}{\textbf{MOA}} & \multicolumn{2}{c||}{\textbf{MOA'}} \\  \cline{2-5}
  \multicolumn{1}{||c||}{$\boldsymbol{n}$} & $\boldsymbol{D^*}$ & $\boldsymbol{f^*}$ & $\boldsymbol{D^*}$ & $\boldsymbol{f^*}$ \\ \hline
  \multicolumn{1}{||c||}{15} & 50 & 267.50 (35.45) & 50 & 202.50 (14.19) \\
  \multicolumn{1}{||c||}{30} & 200 & 1170.00 (94.87) & 100 & 475.00 (42.49) \\
  \multicolumn{1}{||c||}{60} & 800 & 5200.00 (98.46) & 200 & 1050.00 (52.70) \\
  \multicolumn{1}{||c||}{90} & 800 & 5560.00 (126.49) & 400 & 2220.00 (63.25) \\
  \multicolumn{1}{||c||}{120} & 1600 & 11200.00 (871.53) & 800 & 4400.00 (312.33) \\ \hline
  \end{tabular}

\caption{ \small Results of MOA and MOA' (that uses \ibmaphc) 
for the OneMax problem, for increasing problem sizes (rows) 
in terms of critical population size $D^*$, 
and mean and standard deviation over $10$ repetitions 
of the number of fitness evaluations $f^*$ required to obtain the global optimum. 
Lower values of $D^*$ and $f^*$ are better.
\label{tableEDAs1}
}

\end{table*}

We conducted experiments to compare \ibmaphc~as an alternative structure learning within MOA,
denoted $\moaIBMAP$, and denoting by $\moaMI$ the original version that uses MI.
The thesis is that a better structure learning algorithm improves the convergence of MOA, that is,
the optimum is reached computing fewer evaluations of the fitness of individuals.
Both versions were tested on two benchmark functions widely used in the
EDA's literature: \emph{Royal Road} and \emph{OneMax}, both
bit-string optimization tasks, detailed in \cite{Mitchell:1998:IGA:522098}.
The reason these benchmark functions are widely used is that they are hard to optimize,
because the fitness landscape is flat for large areas and then discontinuous.
In the context of evolutionary algorithms these functions model each bit-string as a chromosome
and each bit as a gene.
In the Royal Road problem, the variables are arranged in groups of size $\gamma$.
Its goal is to maximize the number of $1$s in the string, but adding $\gamma$ to the fitness count
only when a group has all $1$s, otherwise adding $0$.
For example, in the case of $\gamma=4$, an individual $111110011111$
is separated in the groups $[1111]~[1001]~[1111]$, and only the first
and third group contribute $4$ to the fitness count, which in
the example equals $8$. The underlying independence structure that should be learned
therefore contains cliques of size $\gamma$, one per group.
In our experiments we used $\gamma=1$ and $\gamma=4$. The former
is known in the literature as \emph{OneMax}. In the example, the fitness
is $10$ for OneMax. Clearly, the optimal individual for both problems is $111111111111$.
%

\begin{table*}[!ht]
\center
  \begin{tabular}{cc|c||c|c||}
   \\ \hline
  \multicolumn{1}{||c||}{} &  \multicolumn{2}{c||}{\textbf{MOA}} & \multicolumn{2}{c||}{\textbf{MOA'}} \\ \cline{2-5}
  \multicolumn{1}{||c||}{$\boldsymbol{n}$} & $\boldsymbol{D^*}$ & $\boldsymbol{f^*}$ & $\boldsymbol{D^*}$ & $\boldsymbol{f^*}$ \\ \hline
  \multicolumn{1}{||c||}{16} & 100 & 545.00 (59.86) & 50 & 337.50 (176.09) \\
  \multicolumn{1}{||c||}{32} & 400 & 3800.00 (210.82) & 400 & 2140.00 (134.99) \\
  \multicolumn{1}{||c||}{64} & 800 & 9120.00 (252.98) & 800 & 4440.00 (126.49) \\
  \multicolumn{1}{||c||}{92} & 1600 & 18400.00 (533.33) & 800 & 5080.00 (500.67) \\
  \multicolumn{1}{||c||}{120} & 1600 & 31120.00 (822.31) & 1600 & 9840.00 (386.44) \\ \hline
  \end{tabular}
  \caption{\small Results of MOA and MOA' (that uses \ibmaphc) 
for the Royal Road problem, for increasing problem sizes (rows) 
in terms of critical population size $D^*$, 
and mean and standard deviation over $10$ repetitions 
of the number of fitness evaluations $f^*$ required to obtain the global optimum.  
Lower values of $D^*$ and $f^*$ are better.
\label{tableEDAs2}}
\end{table*}

In the experiments, MOA is iterated for $1000$ generations
or until the optimum is reached, whatever happened first.
For several runs differing in the initial (random) population,
we measured the \emph{success rate} as the fraction
of times the optimum is found.
A commonly used performance measure in EDAs is the \emph{critical population size}
$D^{*}$; the minimum population size for which
the success rate is $100\%$.
Smaller $D^*$ values have a double benefit on runtime: (i) fewer fitness evaluations for reaching the optima,
and (ii) faster distribution estimation.
We report $D^{*}$ and the number of fitness
evaluations required for that population size, denoted $f^*$.
More robust algorithms are expected to require smaller $D^{*}$ and $f^*$ values.
To measure $D^{*}$ in Royal Road and OneMax,
each version of MOA was run $10$ times for each of
the population sizes $D=\{50,100,200,400,800,1600,3200\}$.
Then, for the measured $D^*$, we report the average and standard deviation
of $f^*$ on each of those runs.
In all the experiments, the population is truncated with a selection size of $50\%$
and an elitism of $50\%$; used for preventing diversity loss.
In $\moaMI$, the parameter $k$ was set to $3$ and $1$ in Royal Road and OneMax, respectively.

%
Results are presented in Table \ref{tableEDAs1} for the OneMax problem, 
and Table \ref{tableEDAs2} for the Royal Road problem.
For both algorithms $\moaMI$ and $\moaIBMAP$, each table reports the values of
$D^*$ as well as both the average and standard deviation of $f^*$, for
increasing problem sizes $n \in \{15,30,60,90,120\}$ for the OneMax problem,
and $n \in \{16,32,64,92,120\}$ for the Royal Road problem (the domain size should be a multiple of $\gamma=4$).
Lower values of $D^*$ and $f^*$ are better.
In both tables, the results show that $\moaIBMAP$ always present equal or lower values of $D^*$ than that of $\moaMI$,
and also  $\moaIBMAP$ always outperforms $\moaMI$ in $f^*$.
For Royal Road, the larger improvement is for $n=92$
where $\moaIBMAP$ requires $75\%$ fewer fitness evaluations $f^*$ and
$D^*$ is halved. For OneMax, the larger improvement is for $n=60$
where $\moaIBMAP$ requires $80\%$ fewer fitness evaluations $f^*$ and
$D^*$ is reduced to a quarter.

An interpretation of these results is that \ibmaphc~estimates better the distribution at each iteration.
To confirm this hypothesis we compared the structures learned by the two algorithms
over our synthetic datasets. For a dataset with $n=75$,
$D=100$, $\tau=2$, the Hamming distances of MI and \ibmaphc~were $132$, and $75$, respectively.
For $\tau=4$ they were $233$ and $143$, respectively; and for $\tau=8$, $395$ and $388$,
respectively. These results show clearly that the quality of \ibmaphc~indeed outperforms
that of MI.
Finally, we highlight that the efficiency of \ibmaphc~allowed it to be run in large
problems up to $120$ genes in size, estimating the structure over many
generations.

\section{Conclusions and future work\label{sec:conclusions}}
This paper proposes IBMAP, a novel independence-based maximum-a-posteriori approach for
learning the structure of Markov networks; and $\ibmaphc$, an
efficient instantiation of IBMAP.
Our approach avoids the cascade errors of traditional independence-based algorithms
that trust completely the outcome of statistical tests.
For this, the central idea of IBMAP is to pose the structure learning task as a maximum-a-posteriori problem, 
by computing the posterior probability of each possible structure given data.
Experiments comparing \ibmaphc~against state-of-the-art independence-based algorithms indicate that our method
improves in most cases over the independence-based competitors with equivalent computational complexities.
\ibmaphc~was also tested in a practical, challenging setting: Estimation of Distribution algorithms,
resulting in faster convergence to the optimum than a state-of-the-art Markov network EDA
algorithm, for the selected benchmark functions.
Our experimental results and the conclusions of Appendix B 
confirm the effectiveness of our structure selection strategy. 
Therefore, we believe that it is worth 
guiding our future work in improving the IB-score as a measure of $\Pr(G \mid D)$,
i.e., relaxing the independence assumption made in Equation~(\ref{eq:independenceApproximationLogs}),
as well as exploring alternative closure sets.
Also, it is clearly worthwhile considering testing our approach in more practical real world testbeds, 
potentially comparing its performance against state-of-the-art score-based algorithms, such as \cite{ganapathi2008,ravikumar2010:l1,DavisAndDomingos2010:BottomUp,vanhaaren2012}.

\section{Acknowledgements}	
This work was funded by the grant PICT-241 of  
the National Agency of Scientific and Technological Promotion, FONCyT, Argentina; 
the grant PID-1205 of the National Technological University, Argentina;
and the scholarship program for teachers of the National Technological University and
the Ministry of Science, Technology and Productive Innovation; Argentina.    
Special thanks to Roberto Santana and Siddhartha Shakya for their help and support while implementing
our experiments on EDAs.

\appendix

\section{Correctness of the Markov blanket closure \label{appendix:mbclosure} }

This appendix presents a formal proof that the Markov blanket closure
described in Definition~\ref{def:mbclosure} of Section~\ref{sec:MB-closure}
is in fact a closure, i.e., its independence assertions completely determine the structure
used to generate it.

Let us start by reproducing some necessary theoretical results
extracted from \cite{koller09,LAURITZEN96,pearl88}: the \emph{pairwise Markov property}, the \emph{Intersection property}
of conditional independence, and the \emph{Strong Union property}
of conditional independence, all satisfied by any Markov network $G$ of a positive graph-isomorph distribution $P$:
\begin{definition}[Pairwise Markov property]
	Let $G$ be a Markov network of some graph-isomorph distribution $P$, then
	\begin{equation}
	(X,Y) \notin E(G) \Leftrightarrow~ \ci{X}{Y}{V\!\setminus\! \{X,Y\}} \mbox{~in $P$}.    \label{eq:pairwise}
    \end{equation}
\end{definition}

\begin{definition}[Intersection]
  The conditional independences among random variables of a positive distribution $P$ satisfy the \emph{Intersection}
  property (expressed in counter-positive form):
  
  \begin{equation} \label{eq:intersection}
		\cd{X}{Y}{\set{Z}} \wedge \ci{X}{W}{\set{Z},Y}  \Rightarrow \cd{X}{Y}{\set{Z},W}
  \end{equation}
  for all $(X \neq Y \neq W) \notin \set{Z}$.
\end{definition}

\begin{definition}[Strong Union]
  The conditional independences among random variables of a graph-isomorph distribution $P$ satisfy the following
  \emph{Strong Union}   property of conditional independence:
  
  \begin{equation} \label{eq:strong-union}
		\ci{X}{Y}{\set{Z}}  \Rightarrow \ci{X}{Y}{\set{Z},W}
  \end{equation}
  for all $(X \neq Y) \notin \set{Z}$.
\end{definition}

We present now two auxiliary lemmas that relate independences with edges in the graph:
 \begin{lemma} \label{lemma:noedge}
	\begin{equation}
		\ci{X}{Y}{\blanket{X}\!\setminus\! \{Y\}}  \Rightarrow~ (X,Y) \notin E(G).     \label{eq:equiv}
	\end{equation}
 \end{lemma}

\textbf{Proof.~}
	The proof proceeds by first applying the Strong union property to the l.h.s. to obtain $\ci{X}{Y}{\set{V}\setminus \{X,Y\}} $,
	and then applying the pairwise property to conclude the r.h.s. $(X,Y) \notin E(G)$.
	\qed


For the remaining of the proof we need to argue that 
something similar to the counter-positive of Lemma~\ref{lemma:noedge} holds:

\begin{lemma} \label{lemma:edge}
	\begin{equation}
		\cd{X}{Y}{\blanket{X}\!\setminus\! \{Y\}}   ~\wedge~ \forall{W  \notin \blanket{X}} \ci{X}{W}{\set{Z},Y} 	 \Rightarrow~ (X,Y) \in E(G).    \label{eq:counter-positive-equiv}
	\end{equation}
\end{lemma}
%
\textbf{Proof.~}
	The proof proceeds by extending the conditioning set $\blanket{X}\!\setminus\! \{Y\}$ of 
	the l.h.s. to the whole domain $V \!\setminus\! \{X,Y\}$, to then apply the counter-positive of \eq{eq:pairwise}
	and reach the r.h.s. $(X,Y) \in E(G)$.  For that, we apply the intersection property of \eq{eq:intersection} 
	iteratively, by taking at each iteration the pair containing 
	one of the independences in
	the l.h.s., and, in the first iteration the  dependence in the l.h.s., 
	and the following iterations the dependence resulting from applying intersection. 
	In all cases, we take $\set{Z} = \blanket{X}\!\setminus\! \{Y\}$. Let see this process in detail.
	In the first iteration we take from the l.h.s. the dependence and the
	independence for the first $W$, obtaining, by intersection, 
	the dependence $\cd{X}{Y}{\set{Z},W}$. We can now take the resulting dependence, with the independence
	for the following $W$, denoted for convenience $W'$. It seems that intersection can no longer be applied 
	 because the respective conditioning sets $\set{Z} \cup \{W\}$ and $\set{Z} \cup \{Y\}$ does not match.
	However, by graph-isomorphism of $P$, we have that the \emph{Strong Union} property of conditional 
	independence is satisfied in $P$, and therefore any independence given some conditioning set follows
	from the same independence given a subset of this conditioning set, in particular then, we have that
	$\ci{X}{W'}{\set{Z},W, Y} $, and intersection can therefore be applied, resulting in $\cd{X}{Y}{\set{Z},W, W'}$.
	Following this iteratively, we reach $\cd{X}{Y}{V \setminus \{X,Y\}}$, where the resulting conditioning
	set $\set{V}\setminus\{X,Y\}$ is the result of $\set{Z} = \blanket{X}\!\setminus\! \{Y\} \cup \blanket{X}$, recalling $X \notin \blanket{X}$.

	\qed


We can now prove our main theorem:
\newtheorem{thma}{Theorem}
\begin{thma} 
	Let $G$ be an undirected independence structure of a positive graph-isomorph distribution $P(\set{V})$.
	The \emph{Markov blanket closure} of $G$
	is a set of conditional independence assertions 
	that are sufficient for completely determining the structure $G$. 
\end{thma}
\textbf{Proof.~}
We prove the above theorem by proving
that all the edges and no edges in $G$ are determined by the assertions
contained in $\closure(G)$. 
We do it separately for absence and existence of edge between any two variables $X$ and $Y$:

\begin{itemize}
 \item[i)] \textbf{For edge absence:} 
Let $(X,Y) \notin E(G)$. Then, by definition, the closure contains the two independence assertions: 
$\ci{X}{Y}{\blanket{X} \!\setminus\! \{Y\} }$ and $\ci{Y}{X}{\blanket{Y} \!\setminus\! \{X\} }$, which,
by \eq{eq:equiv} of Lemma~\ref{lemma:noedge} both imply $(X,Y) \notin E(G)$.
 
\item[ii)]\textbf{For edge existence:} \\
Similarly, let
$(X,Y) \in E(G)$. Then, by definition, the closure contains the dependence assertion: 
$\cd{X}{Y}{\blanket{X} \!\setminus\! \{Y\} }$.
Also, for all $W$ s.t. $(X,W) \notin E(G)$ (i.e., $W \notin \blanket{X}$), the closure contains $\ci{X}{W}{\blanket{X} }$.
Then, by \eq{eq:counter-positive-equiv} of Lemma~\ref{lemma:edge} we have that $(X,Y) \in E(G)$.
 \qed
\end{itemize}

\section{IBMAP landscape analysis \label{appendix:landscape}}

In this appendix we report the results of an experiment that analyzes empirically
the landscape of the IB-score function on synthetic datasets.
The experiment consists in an analysis of the surface of the IB-score 
over the complete search space of possible structures.
The aim is to assess how good is the hill-climbing search for maximizing the IB-score. 
Due to the exponential number of possible networks for each domain,
in a first instance we explore how the complete landscape of IB-score looks like
for datasets with a small domain size $n=6$.
For this experiment, we used synthetic datasets similar to those used in Section~\ref{sec:randomexperiments}.
\begin{figure}[!ht]
\center
\scriptsize
$~~~~~~~n~=~6$
\normalsize
 \includegraphics[width=1\textwidth]{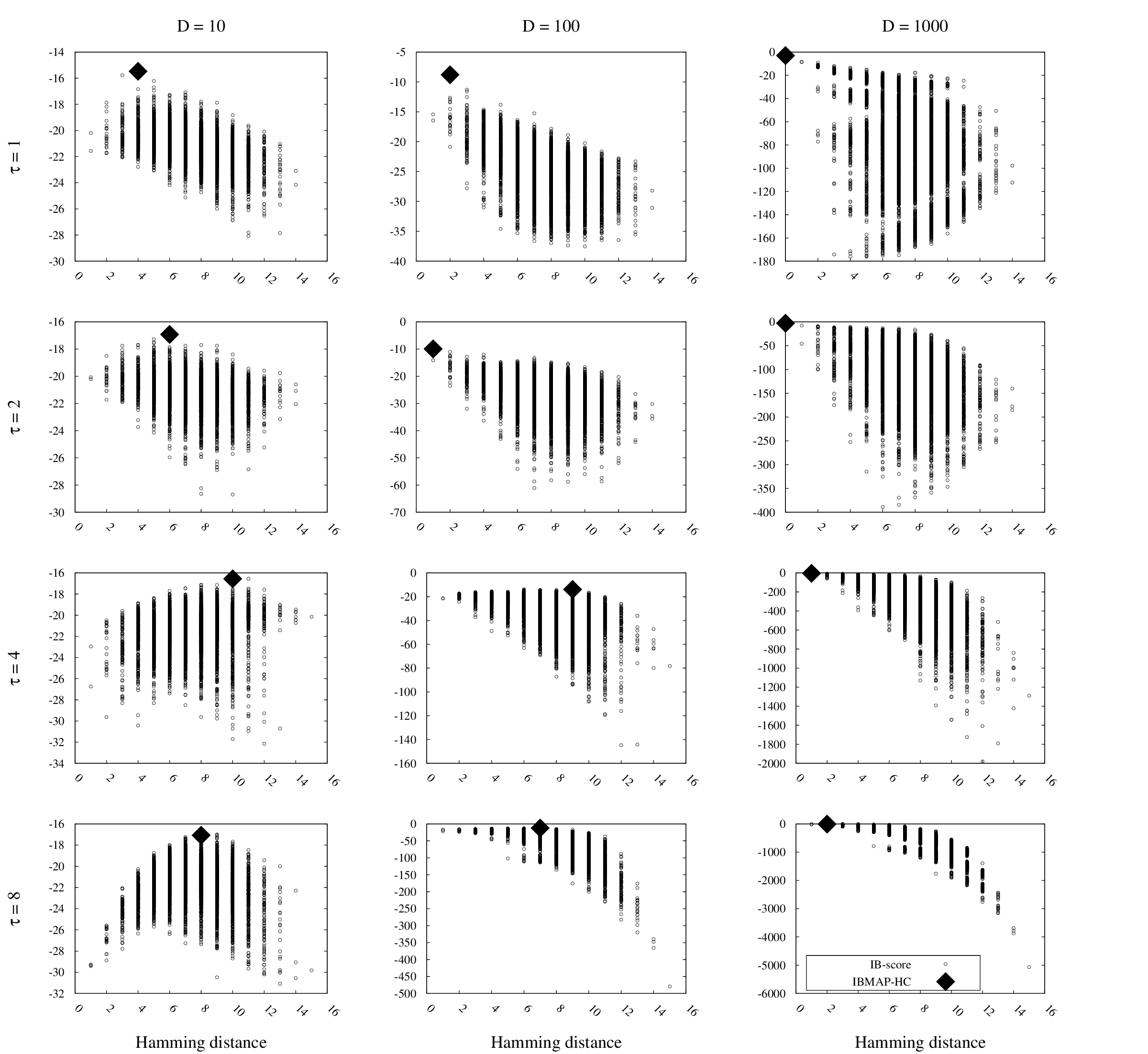} 
\caption{\small Complete landscape of the IB-score for synthetic datasets with $n=6$, 
for increasing dataset sizes $D=10$ (first column), $D=100$ (second column), and
  $n=1000$ (third column), and $\tau \in \{1,2,4,8\}$ in the rows. 
The X-axis sort the structures in the Hamming distance with the correct structure.
The Y-axis shows the IB-score for all the structures in the landscape.
The structure found by \ibmaphc~is indicated by a diamond. \label{fig:landscape_n6} }
\end{figure}

The plots in Figure~\ref{fig:landscape_n6} 
show in the Y-axes the values of the IB-score for all the possible structures,
and sort the structures in the X-axes, by its Hamming distance 
to the true underlying structure in the dataset 
(this is, from zero, to $n \choose 2$). 
Note that the scores of the structures appear in log probabilities, 
because they was computed as shown in Equation~(\ref{eq:independenceApproximationLogs}).
With this layout, the structures in the left (near to zero) 
are those with less structural errors, 
and are also those expected to have a higher value of the IB-score. 
Therefore, the structures in the right are expected to have lower values of the IB-score. 
Also, indicated with a diamond, the structures found by the algorithm \ibmaphc~are shown for each case.

\begin{figure}[!ht]
\center
\scriptsize
$~~~~~~~~n~=~20$
\normalsize

 \includegraphics[width=1\textwidth]{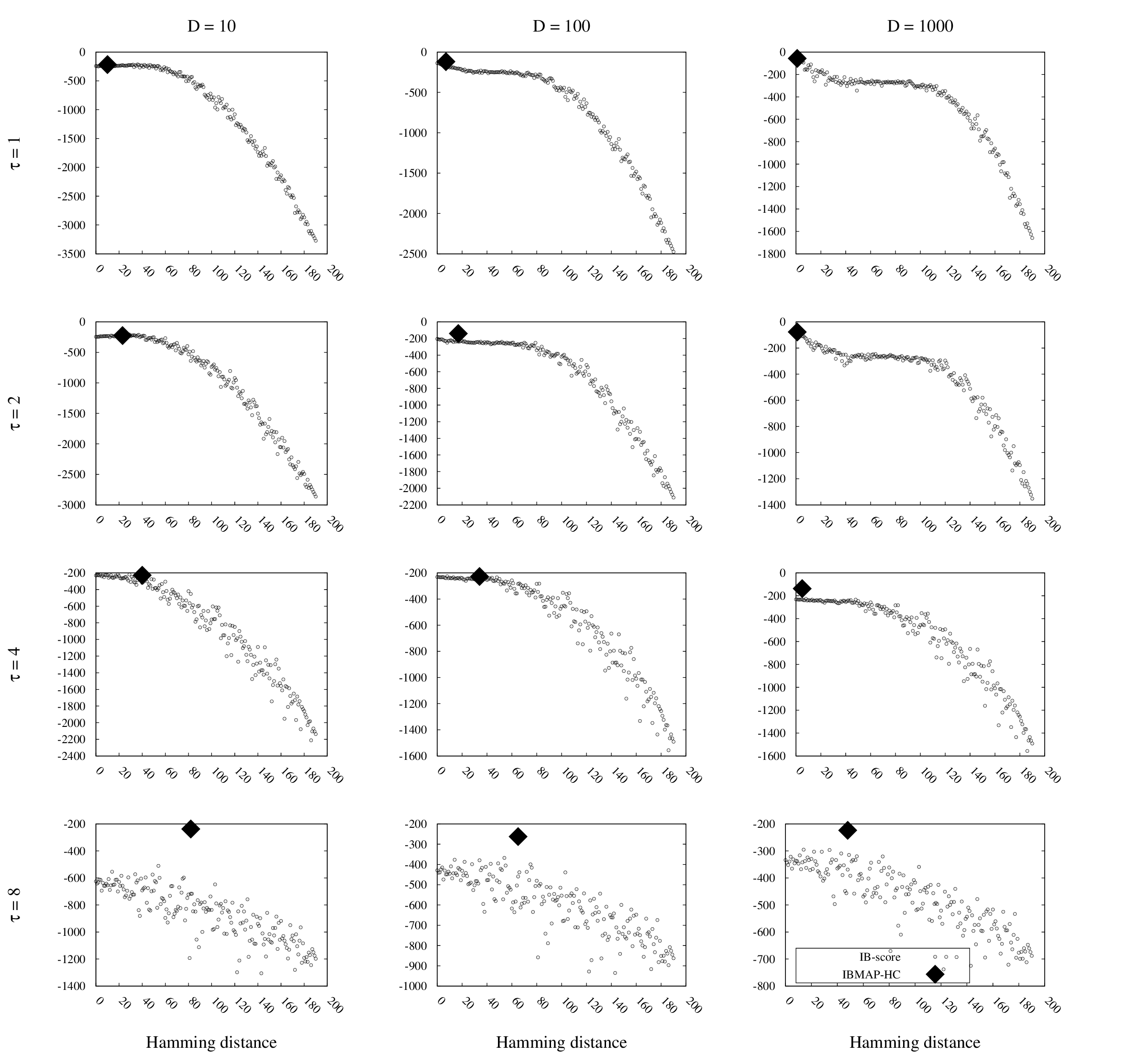} 
\caption{\small A fraction of the landscape of the IB-score for synthetic datasets with $n=20$, 
for increasing dataset sizes $D=10$ (first column), $D=100$ (second column), and
  $n=1000$ (third column), and $\tau \in \{1,2,4,8\}$ in the rows. 
The X-axis sort the structures in the Hamming distance with the correct structure.
The Y-axis shows the IB-score for all the structures in the landscape.
The structure found by \ibmaphc~is indicated by a diamond. \label{fig:landscape_n20} }
\end{figure}

The plots are ordered in the columns for increasing values 
of the dataset $D\in\{10,100,1000\}$, and in the rows, 
the different values of $\tau\in\{1,2,4,8\}$, 
increasing the complexity of the problem.
From the analysis of such plots, it is observed 
how the landscape shapes to a decreasing curve as increasing the value $D$ 
(see the tendency from left to right columns, and not the change in scale in the Y-axis). 
This is achieved because the precision of the statistical tests 
improves with increasing $D$.
In second place, the diamond that indicates the position in the landscape 
of the structure learned by the \ibmaphc~algorithm, 
achieves always the structure with highest score value.
It can be also observed how the error of the structure 
learned by \ibmaphc~is closer to zero while increasing $D$.

A second instance of this experiment was made for a domain size $n=20$.
In this instance, the landscape contains a total size of $2^{20 \choose 2}$. 
As it is impossible to show the IB-score for the complete landscape,
we show only a subset obtained by generating 
randomly $5$ structures deferring in $m$ edges to the true structure, 
with $m$ from $0$ to $20 \choose 2$ in the X-axis.
Such results are shown in Figure~\ref{fig:landscape_n20}.
From the analysis of such plots, the same conclusions are observed.

To conclude this appendix, it is worth noting that our results confirm 
the effectiveness of our structure selection strategy in maximizing the IB-score over the complete landscape.
For that reason, we conclude that it is worth guiding our future work 
only in the improvement of the IB-score as a measure of $\Pr(G \mid D)$.

\bibliographystyle{spmpsci}

\begin{thebibliography}{10}
\providecommand{\url}[1]{{#1}}
\providecommand{\urlprefix}{URL }
\expandafter\ifx\csname urlstyle\endcsname\relax
  \providecommand{\doi}[1]{DOI~\discretionary{}{}{}#1}\else
  \providecommand{\doi}{DOI~\discretionary{}{}{}\begingroup
  \urlstyle{rm}\Url}\fi

\bibitem{Asuncion+Newman:2007}
A.~Asuncion, D.N.: {UCI} machine learning repository (2007)

\bibitem{AGRESTI02}
Agresti, A.: {Categorical Data Analysis}, 2nd edn.
\newblock Wiley (2002)

\bibitem{Alden07}
Alden, M.: {MARLEDA: Effective Distribution Estimation Through Markov Random
  Fields}.
\newblock Ph.D. thesis, Dept of CS, University of Texas Austin (2007)

\bibitem{Aliferis2010}
Aliferis, C., Statnikov, A., Tsamardinos, I., Mani, S., Koutsoukos, X.: {Local
  Causal and Markov Blanket Induction for Causal Discovery and Feature
  Selection for Classification Part I: Algorithms and Empirical Evaluation}.
\newblock JMLR \textbf{11}, 171--234 (2010)

\bibitem{Aliferis2010b}
Aliferis, C., Statnikov, A., Tsamardinos, I., Mani, S., Koutsoukos, X.: {Local
  Causal and Markov Blanket Induction for Causal Discovery and Feature
  Selection for Classification Part II: Analysis and Extensions}.
\newblock JMLR \textbf{11}, 235--284 (2010)

\bibitem{Aliferis2003:HITON}
Aliferis, C., Tsamardinos, I., Statnikov, A.: {HITON}, a novel {Markov} blanket
  algorithm for optimal variable selection.
\newblock AMIA Fall  (2003)

\bibitem{BrombMarg09}
Bromberg, F., Margaritis, D.: {Improving the Reliability of Causal Discovery
  from Small Data Sets using Argumentation}.
\newblock JMLR \textbf{10}, 301--340 (2009)

\bibitem{Bromberg06}
Bromberg, F., Margaritis, D., Honavar, V.: {Efficient markov network structure
  discovery using independence tests}.
\newblock In: In Proc SIAM Data Mining, p.~06 (2006)

\bibitem{brombergmargaritis09b}
Bromberg, F., Margaritis, D., V., H.: {Efficient Markov Network Structure
  Discovery Using Independence Tests}.
\newblock JAIR \textbf{35}, 449--485 (2009)

\bibitem{Chickering96lns}
Chickering, D.M.: Learning {Bayesian} networks is {NP}-{Complete}.
\newblock In: D.~Fisher, H.~Lenz (eds.) Learning from Data: Artificial
  Intelligence and Statistics V, pp. 121--130. Springer-Verlag (1996)

\bibitem{covertomas91}
Cover, T.M., Thomas, J.A.: {Elements of information theory}.
\newblock Wiley-Interscience, New York, NY, USA (1991)

\bibitem{cressie92}
Cressie, N.: Statistics for spatial data. 
\newblock Terra Nova 4(5):613--617,
\newblock \doi{10.1111/j.1365-3121.1992.tb00605.x}
  
\bibitem{DavisAndDomingos2010:BottomUp}
Davis, J., Domingos, P.: {Bottom-Up Learning of Markov Network Structure}.
\newblock In: ICML, pp. 271--278 (2010)

\bibitem{PietraPL97}
Della~Pietra, S., Della~Pietra, V.J., Lafferty, J.D.: {Inducing Features of
  Random Fields}.
\newblock IEEE Trans. PAMI. \textbf{19}(4), 380--393 (1997)

\bibitem{friedman00}
Friedman, N., Linial, M., Nachman, I., Pe'er, D.: {Using {Bayesian} Networks to
  Analyze Expression Data}. 
  \newblock Journal of computational biology, pp. 601--620 (2000)
  
\bibitem{ganapathi2008}
Ganapathi, V., Vickrey, D., Duchi, J., Koller, D.: {Constrained Approximate
  Maximum Entropy Learning of Markov Random Fields}.
\newblock In: Uncertainty in Artificial Intelligence, pp. 196--203 (2008)

\bibitem{Hammersley_Clifford_1968}
Hammersley, J. M., Clifford, P.: Markov fields on finite graphs and lattices (1968).

\bibitem{Hettich+Bay:1999}
Hettich, S., Bay, S.D.: The {UCI KDD} archive (1999)

\bibitem{koller09}
Koller, D., Friedman, N.: {Probabilistic Graphical Models: Principles and
  Techniques}.
\newblock MIT Press (2009)

\bibitem{larranagalozano2002}
Larra{\~{n}}aga, P., Lozano, J.A.: {E}stimation of {D}istribution {A}lgorithms.
  {A} {N}ew {T}ool for {E}volutionary {C}omputation.
\newblock Kluwer Pubs (2002)

\bibitem{LAURITZEN96}
Lauritzen, S.L.: Graphical Models.
\newblock Oxford University Press (1996)

\bibitem{Lee+al:NIPS06}
Lee, S.I., Ganapathi, V., Koller, D.: Efficient structure learning of {M}arkov
  networks using {L1}-regularization.
\newblock In: NIPS (2006)

\bibitem{Li2009}
Li, S.: {Markov random field modeling in image analysis}. 
  \newblock Springer, 2009.

\bibitem{MARGARITIS05}
Margaritis, D.: {Distribution-Free Learning of Bayesian Network Structure in
  Continuous Domains}.
\newblock In: Proceedings of AAAI (2005)

\bibitem{margaritisBromberg09}
Margaritis, D., Bromberg, F.: {Efficient Markov Network Discovery Using
  Particle Filter}.
\newblock Comp. Intel. \textbf{25}(4), 367--394 (2009)

\bibitem{MARGARITIS00}
Margaritis, D., Thrun, S.: {Bayesian} network induction via local
  neighborhoods. 
 \newblock In: Proceedings of NIPS06 (2000) 
  
\bibitem{MCCALLUM03}
McCallum, A.: Efficiently inducing features of conditional random fields.
\newblock In: Proceedings of Uncertainty in Artificial Intelligence (UAI)
  (2003)

\bibitem{minka2005}
Minka, T.: {Divergence measures and message passing}.
\newblock Tech. rep., Microsoft Research (2005)

\bibitem{Mitchell:1998:IGA:522098}
Mitchell, M.: {An Introduction to Genetic Algorithms}.
\newblock MIT Press, Cambridge, MA, USA (1998)

\bibitem{Muhlenbein96}
M{\"u}hlenbein, H., Paa{\ss}, G.: From recombination of genes to the estimation
  of distributions {I}. binary parameters.
\newblock In: H.M. Voigt, W.~Ebeling, I.~Rechenberg, H.P. Schwefel (eds.)
  Parallel Problem Solving from Nature — PPSN IV, \emph{Lecture Notes in
  Computer Science}, vol. 1141, pp. 178--187. Springer Berlin / Heidelberg
  (1996).
\newblock 10.1007/3-540-61723-X\_982

\bibitem{pearl88}
Pearl, J.: {Probabilistic Reasoning in Intelligent Systems: Networks of
  Plausible Inference}.
\newblock Morgan Kaufmann Publishers, Inc. (1988)

\bibitem{ravikumar2010:l1}
Ravikumar, P., Wainwright, M.J., Lafferty, J.D.: {High-dimensional Ising model
  selection using L1-regularized logistic regression}.
\newblock Annals of Statistics \textbf{38}, 1287--1319 (2010).
\newblock \doi{10.1214/09-AOS691}

\bibitem{Santana:2005:kikuchi}
Santana, R.: Estimation of distribution algorithms with kikuchi approximations.
\newblock Evol. Comput. \textbf{13}(1), 67--97 (2005).
\newblock \doi{10.1162/1063656053583496}.
\newblock \urlprefix\url{http://dx.doi.org/10.1162/1063656053583496}

\bibitem{schluter2012survey}
Schl{\"u}ter, F.: A survey on independence-based markov networks learning.
\newblock Artificial Intelligence Review pp. 1--25 (2012).
\newblock \urlprefix\url{http://dx.doi.org/10.1007/s10462-012-9346-y}.
\newblock 10.1007/s10462-012-9346-y

\bibitem{Shakya_McCall_2007}
Shakya, S., McCall, J.: Optimization by estimation of distribution with deum
  framework based on markov random fields.
\newblock International Journal of Automation and Computing \textbf{4}(3),
  262--272 (2007).
\newblock
  \urlprefix\url{http://www.springerlink.com/index/10.1007/s11633-007-0262-6}

\bibitem{moapaper}
Shakya, S., Santana, R., Lozano, J.A.: A markovianity based optimisation
  algorithm.
\newblock Genetic Programming and Evolvable Machines \textbf{13}(2), 159--195
  (2012)

\bibitem{shekhar04}
Shekhar, S., Zhang, P., Huang, Y., Vatsavai, R. R.: 
Trends in spatial data mining. Data mining: Next generation challenges and future directions, 357-380 (2003).
  
\bibitem{schmidts08}
Schmidt, M., Murphy, K., Fung, G., Rosales, R.: Structure learning in random
  fields for heart motion abnormality detection. 
  \newblock In: Computer Vision and
  Pattern Recognition, 2008. CVPR 2008. IEEE Conference on, pp 1 --8,
  \doi{10.1109/CVPR.2008.4587367}
  
\bibitem{Spirtes00}
Spirtes, P., Glymour, C., Scheines, R.: {Causation, Prediction, and Search}.
\newblock Adaptive Computation and Machine Learning Series. MIT Press (2000)

\bibitem{vanhaaren2012}
Van~Haaren, J., Davis, J.: Markov network structure learning: {A} randomized
  feature generation approach.
\newblock In: Proceedings of the Twenty-Sixth AAAI Conference on Artificial
  Intelligence (2012)
\newblock \urlprefix\url{https://lirias.kuleuven.be/handle/123456789/345604}

\bibitem{vanhaaren2013}
Van~Haaren, J., Davis, J., Lappenschaar, M., Hommersom, A.: 
Exploring disease interactions using {M}arkov networks. 
\newblock In:  Proceedings of the AAAI-2013 (HIAI-2013). 
Bellevue, Washington, United States, 15 July, (2013)

\bibitem{wainwright2008}
Wainwright, M. J., Jordan, M. I: Graphical models, exponential families, and variational inference. 
Foundations and Trends in Machine Learning, 1(1-2), 1-305. (2008)

\bibitem{Welsh:1993:CKC:164784}
Welsh, D.J.A.: Complexity: knots, colourings and counting.
\newblock Cambridge University Press, New York, NY, USA (1993)


\end{thebibliography}

\end{document}